\documentclass[journal]{IEEEtran}
\IEEEoverridecommandlockouts
\usepackage{cite}
\usepackage{amsmath,amssymb,amsfonts}
\usepackage{algorithmic}
\usepackage{graphicx}
\usepackage{textcomp}
\usepackage{xcolor}
\usepackage{pgfplots}
\usepackage{pgfplotstable}
\usepackage{array}
\usepackage{booktabs}
\usepackage{ulem}
\usepackage{hyperref}
\pgfplotsset{compat=1.17}
\usepackage{tikz}
\usepackage{adjustbox}
\usepackage{enumitem}
\usepackage{arydshln}
\usepackage{xstring}
\renewcommand{\arraystretch}{1.8}
\usepackage{multirow}
\usepackage[caption=false,font=normalsize,labelfont=sf,textfont=sf]{subfig}
\usepackage{stfloats}
\usepackage{url}
\usepackage{verbatim}
\usepackage{fontenc}
\usepackage{pifont}
\usepackage{tabularx}
\usepackage{threeparttable}
\usepackage[edges]{forest}
\usetikzlibrary{shadows}
\usetikzlibrary{shadows.blur}
\usetikzlibrary{trees, positioning}
\definecolor{rootgreen}{RGB}{215, 235, 215}
\definecolor{catpink}{RGB}{245, 210, 210}
\definecolor{suborange}{RGB}{255, 235, 205}
\definecolor{leafblue}{RGB}{230, 240, 250}

\begin{document}
	
	\title{Survey and Experiments on Mental Disorder Detection via Social Media: From Large Language Models and RAG to Agents}
	\author{
		Zhuohan Ge, Darian Li, Yubo Wang, Nicole Hu, Xinyi Zhu, Haoyang Li, Xin Zhang, Mingtao Zhang, Shihao Qi, Yuming Xu, Han Shi, Chen Jason Zhang, and Qing Li,~\IEEEmembership{Fellow,~IEEE}
	\thanks{Zhuohan Ge, Darian Li, Nicole Hu, Haoyang Li, Mingtao Zhang, Shihao Qi, Yuming Xu, Chen Jason Zhang and Qing Li are from The Hong Kong Polytechnic University, Hong Kong SAR, China.}
	\thanks{Yubo Wang, Xinyi Zhu, Xin Zhang, Han Shi are from The Hong Kong University of Science and Technology, Hong Kong SAR, China.}}

\markboth{IEEE Transactions on Knowledge and Data Engineering,~Vol.~XX, No.~XX, XX~2025}%
{\MakeLowercase{\textit{Ge et al.}}: A Survey of Large Language Models in Mental Health Disorder Detection on Social Media}

	\maketitle
	
	\begin{abstract}
		Mental disorders represent a critical global health challenge, and social media is increasingly viewed as a vital resource for real-time digital phenotyping and intervention.
		To leverage this data, large language models (LLMs) have been introduced, offering stronger semantic understanding and reasoning than traditional deep learning, thereby enhancing the explainability of detection results.
		Despite the growing prominence of LLMs in this field, there is a scarcity of scholarly works that systematically synthesize how advanced enhancement techniques, specifically Retrieval-Augmented Generation (RAG) and Agentic systems, can be utilized to address these reliability and reasoning limitations.
		Here, we systematically survey the evolving landscape of LLM-based methods for social media mental disorder analysis, spanning standard pre-trained language models, RAG to mitigate hallucinations and contextual gaps, and agentic systems for autonomous reasoning and multi-step intervention. 
		We organize existing work by technical paradigm and clinical target, extending beyond common internalizing disorders to include psychotic disorders and externalizing behaviors.
		Additionally, the paper comprehensively evaluates the performance of LLMs, including the impact of RAG, across various tasks. 
		This work establishes a unified benchmark for the field, paving the way for the development of trustworthy, autonomous AI systems that can deliver precise and explainable mental health support.
	\end{abstract}
	

	\section{Introduction}
	\IEEEPARstart{G}{lobally}, half of all individuals will experience or have experienced a mental health disorder \cite{mcgrath2023age}, and mental health issues have become a significant challenge affecting the well-being of both societies and individuals. 
	According to the World Health Organization (WHO) \cite{WHO}, in 2021, nearly 1.1 billion people worldwide were affected by a mental disorder, representing 14.3\% of the global population \cite{who-mental-disorders}. Figure \ref{fig:fig1} provides a detailed breakdown of these statistics, revealing that anxiety and depression affect approximately 359 million and 280 million individuals, respectively.
	Despite the severity of the problem, traditional mental disorder detection methods, such as questionnaires\cite{van2016validation,alfonsson2014interformat}, are limited in early, large-scale application by their lag behind real-time events and their reliance on self-reporting, which is significantly compromised by recall and social desirability biases\cite{latkin2017relationship}. 

	\begin{figure}[t]
    \centering
    \begin{tikzpicture}
        \begin{axis}[
            ybar,
            bar width=0.5cm,
			enlarge x limits=0.25,
			bar shift=0pt,
            width=8cm,
            height=6cm,
            ymin=0,
            ymax=450,
            ylabel={Population (Millions)},
            xlabel style={font=\small}, 
            ylabel style={font=\small},
            symbolic x coords={Dep., Anx., Bip., Sch., Con.},
			xtick={Dep.,Anx.,Bip.,Sch.,Con.},
			xticklabels={Dep.,Anx.,Bip.,Sch.,Con.},
            nodes near coords,
            nodes near coords align={vertical},
			xtick pos=bottom,
            ytick pos=left,
        ]
        \definecolor{DepColor}{rgb}{0.93, 0.70, 0.25}
        \definecolor{AnxColor}{rgb}{0.89, 0.42, 0.36}
        \definecolor{BipColor}{rgb}{0.58, 0.46, 0.75}
        \definecolor{SchColor}{rgb}{0.45, 0.73, 0.38}
        \definecolor{ConColor}{rgb}{0.27, 0.48, 0.85}

        \addplot[ybar, fill=DepColor] coordinates {(Dep., 280)};
        \addplot[ybar, fill=AnxColor] coordinates {(Anx., 359)};
        \addplot[ybar, fill=BipColor] coordinates {(Bip., 37)};
        \addplot[ybar, fill=SchColor] coordinates {(Sch., 23)};
        \addplot[ybar, fill=ConColor] coordinates {(Con., 41)};
        \end{axis}
    	\end{tikzpicture}
    	\vspace{-10px}
        \caption{Global Population with Mental Disorders (2021) \cite{who-mental-disorders}, where ‘Dep.’ stands for depression, ‘Anx.’ for anxiety, ‘Bip.’ for bipolar, ‘Sch.’ for schizophrenia, and ‘Con.’ for conduct-dissocial disorder.}
		\vspace{-10px}
		\label{fig:fig1}
	\end{figure}

	As a result, social media has emerged as an important source~\cite{naslund2020social,chen2021social}, providing large-scale, real-time and relatively spontaneous expressions that reflects not only users' daily lives but also their emotional fluctuations and psychological states, which are less susceptible to the reporting biases found in traditional questionnaires.
	With an immense active user base \cite{social-app-report}, these platforms generate massive data that capture both daily behaviors and emotional and psychological signals.
	According to Zhang et al.~\cite{zhang2022natural}, among the 399 studies included in their review, social media text dominates as the primary data source (81\%), far exceeding interviews (7\%) and electronic health records (6\%).

	However, social media data is inherently unstructured and characterized by semantic complexity and label sparsity \cite{abkenar2021big,gao2022modeling}. 
	Effectively extracting reliable mental health signals from such high-entropy environments necessitates systems capable of deep semantic interpretation and complex reasoning. 
	Without these capabilities, detection models remain dependent on extensive labeled datasets and capture only surface-level statistical correlations, often missing the nuanced psychological intent hidden within the noise.

	While traditional machine learning, deep learning, and early pre-trained language models (e.g., BERT) laid the foundation for text analysis, they often fall short in meeting these reasoning and interpretability demands. 
	With the rapid development of natural language processing, large language models (LLMs)~\cite{hadi2023survey,li2024survey} have bridged this gap. 
	With their pre-trained world knowledge and superior reasoning capabilities, LLMs can effectively interpret these subtle mental health signals even in low-resource settings.
	These models transcend simple classification by offering a deep understanding of syntactic and semantic contexts \cite{mudrik2024exploring,omar2024applications}.
	They enable not only the detection of mental disorder symptoms from user posts but also the generation of diagnostic explanations based on rich pre-trained knowledge. Despite these advancements, applying LLMs in high-stakes domains remains challenging.
	Without specialized medical grounding, LLMs can suffer from ‘hallucinations', generating unfaithful or fabricated responses \cite{zhang2023large,huang2025survey}, which is unacceptable for mental disorder detection.
	Moreover, LLMs lack persistent memory and explicit planning, making them insufficient for executing complex, multi-step clinical decision workflows.

	To address these limitations, Retrieval-Augmented Generation (RAG) has emerged as a powerful enhancement to LLMs. 
	It integrates retrieval mechanisms that apply a fixed retrieval strategy to fetch relevant external information as supporting context to aid LLMs' generation.
	Based on the standard RAG, LLM-based agentic systems further introduce extra tools and memory modules. 
	These extra modules make the models able to plan for their retrieval tailored for each query, as well as better organize their evolving memory to reduce the computational costs.
	Consequently, they can augment LLMs for mental disorder diagnosis and enable them to handle complex, multi-step mental health intervention tasks autonomously.

	While existing surveys \cite{wongkoblap2017researching,skaik2020using,harrigian2020state,lejeune2022use,malhotra2022deep,garg2023mental,di2023methodologies,owen2024ai,shah2024mental,omar2024exploring,hua2024large,hua2024applying,ke2024exploring,guo2024large,lawrence2024opportunities} have explored different approaches and challenges in detecting mental disorders, a significant gap remains: none have investigated the role of RAG techniques and agentic systems.
	This oversight leaves unexplored how RAG integration enhances factual reliability through external knowledge retrieval, and how agentic systems introduce a paradigm shift towards autonomous reasoning and persistent memory, enabling the orchestration of complex diagnostic workflows in the noisy social media environment.
	Additionally, many current surveys \cite{hua2024large, hua2024applying, ke2024exploring, guo2024large, lawrence2024opportunities} primarily focus on the general application of LLMs in the mental health field, with less emphasis placed on social media contexts, where individuals frequently express mental disorder symptoms \cite{naslund2020social}.
	Neglecting social media data, which provides real-time insights for early intervention and long-term data to model disease progression, limits our understanding of how LLMs can be effectively applied to these dynamic platforms.

	Specifically, this survey provides a comprehensive overview of the current state of the art in the application of LLMs for detecting various mental disorders on social media, as well as how RAG techniques and agentic systems enhance their capabilities.
	We also offer a unified benchmarking and empirical analysis across multi-disorder and multi-task settings. Our contributions are summarized as follows:
	\begin{itemize}[leftmargin=*]
	\item We discuss a wide range of mental disorders, elaborating on their features, differences, and commonalities to enhance data mining and computational modeling.
	\item We delineate some specific applications of LLMs on social media and their key research methods, models, datasets, etc. 
	\item We conduct the first survey on the RAG techniques and agentic systems tailored for mental disorder detection.
	\item We present popular social media benchmark datasets, highlighting their characteristics and applicable tasks. 
	\item We conduct extended experiments that not only evaluate the capabilities of LLMs under different diseases but also assess the impact of RAG techniques for LLMs.
	\end{itemize}

	\textbf{Organization.} The remainder of this paper is structured as follows.
	Section \ref{sec:2} introduces the background and concepts of mental disorders, social media platforms, LLMs, RAG techniques and agentic systems.
	Section \ref{sec:3} surveys existing literature across key paradigms, including DLMs, PLMs, standard LLMs, and hybrid approaches incorporating RAG and agentic systems, followed by a discussion on future directions in areas where research is lacking.
	Section \ref{sec:4} explores popular datasets and evaluation metrics.
	Section \ref{sec:5} evaluates the performance of multiple LLMs on different tasks and the impact of RAG methods on LLMs.
	Section \ref{sec:6} outlines the limitations, challenges, and future directions.
	Finally, Section \ref{sec:7} concludes with key insights and implications.

		\section{PRELIMINARY} \label{sec:2}
	In this section, we provide a taxonomy of common mental disorders and various social media data types. Then we explain why LLMs, RAG, and agentic systems are particularly advantageous.

	\subsection{Main Mental Disorders} \label{subsec:disor}
	Mental disorders not only affect an individual's mood and behavior but also significantly impact daily life and social functioning, and can even lead to self-harm and suicidal tendencies \cite{hall2019association,chun2025gambling}, emphasizing the importance of early detection and intervention\cite{costello2016early}.

	Based on the broad diagnostic categories defined in previous studies \cite{garabiles2019exploring,caspi2020longitudinal,cicchetti2014developmental}, mental disorders can be generally classified into internalizing disorders (e.g., depression, anxiety) and externalizing disorders (e.g., conduct disorder, ADHD).
	However, an analysis of the classification changes of psychotic disorders identified substantial heterogeneity in the symptoms and disease course of schizophrenia and other mental disorders, suggesting that schizophrenia constitutes a distinct spectrum independent of the internalizing-externalizing framework due to its unique comorbidity structure \cite{kotov2011schizophrenia,biedermann2016psychotic}.
	Therefore, we additionally incorporate psychotic (e.g., schizophrenia) as a separate category, ultimately classifying mental disorders into three distinct groups: internalizing disorders, psychotic disorders, and externalizing disorders.

\begin{figure*}[t]
		\centering
		\includegraphics[width=0.95\textwidth]{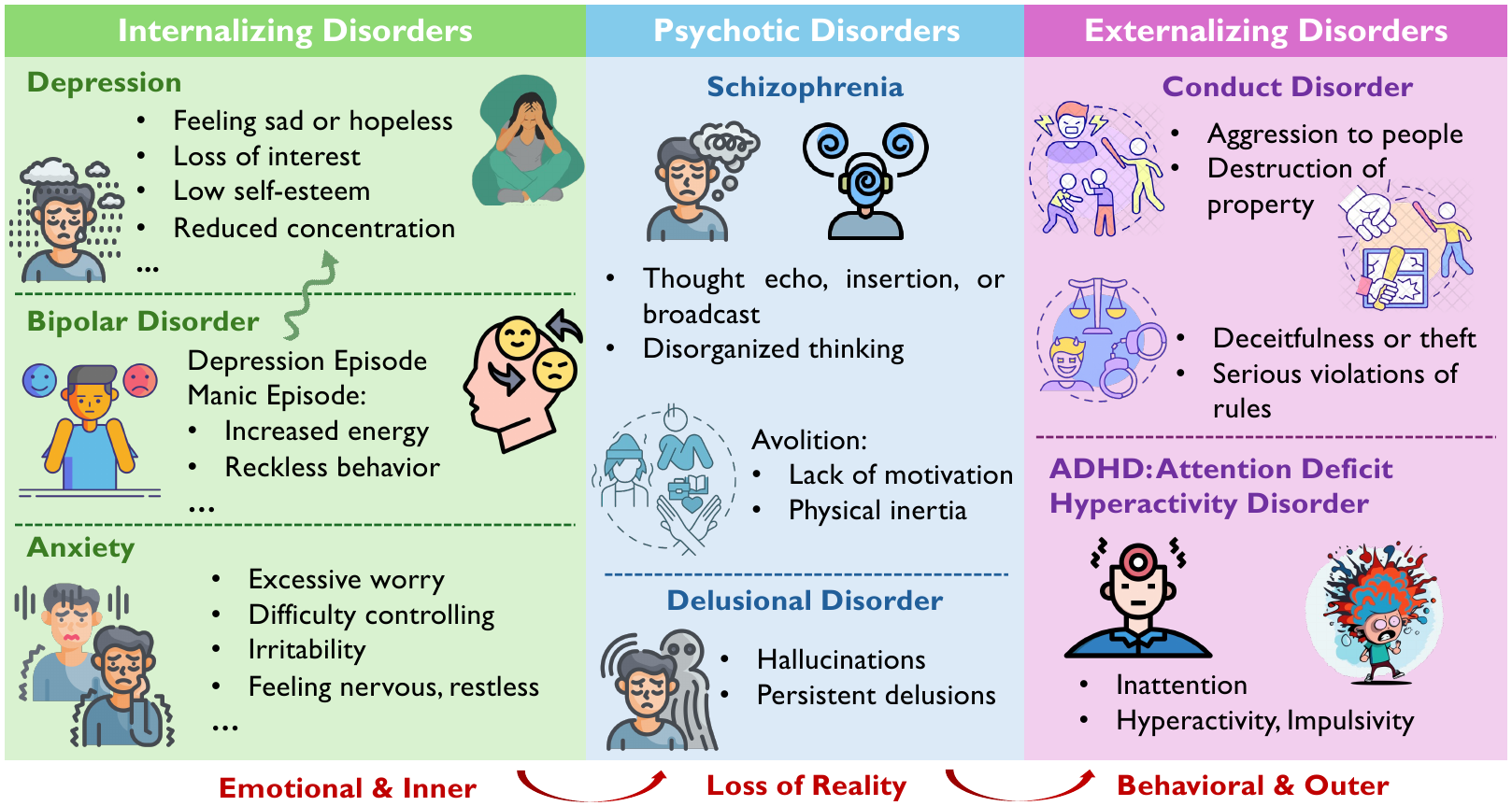}
		\caption{Taxonomy and Symptom Introduction of Mental Disorders.}
		\label{fig:fig2}
		\vspace{-10px}
	\end{figure*}

	Next, we will elaborate on the main features of these three types of mental disorders. Figure \ref{fig:fig2} presents the specific symptoms and categorizations of the primary diseases associated with these mental disorders.
	
	\begin{itemize}[leftmargin=*]
		\item \textbf{Internalizing Disorders} are characterized by internal distress, primarily manifested through persistent negative emotions, excessive worry, and a pervasive sense of hopelessness toward life\cite{myers2002ten}. 
		Individuals with these disorders typically remain cognitively aware of their struggles, presenting without psychotic symptoms and with intact reality testing ability, with their distress primarily centered on self-directed suffering rather than outward disruption.
		Moreover, comorbidity between internalizing disorders may occur in certain cases \cite{mcconaughy1993comorbidity,garabiles2019exploring}.
		
		\item \textbf{Psychotic Disorders} are characterized by perceptual disturbances and a fundamental break from reality, often manifesting as hallucinations, delusions, and disorganized thinking\cite{lieberman2018psychotic}. 
		These conditions distort an individual's perception of the world, resulting in profound cognitive disruptions and difficulties in maintaining logical thought processes, which ultimately lead to significant impairments in perception, communication, and daily social functioning.
		
		\item \textbf{Externalizing Disorders} encompass a spectrum of conditions, including disruptive behavior and dissocial disorders, which are primarily characterized by impulsivity, aggression, and rule-breaking behaviors. 
		Conditions such as conduct disorder and attention deficit hyperactivity disorder (ADHD) often lead individuals to engage in disruptive actions, typically manifesting as hostility, defiance, or difficulties in self-regulation. 
		These overtly disruptive behaviors often lead to dysfunctional behavioral control and conflicts with peers or authority figures.
	\end{itemize}

	\subsection{Social Media Data Types}
\begin{figure}[t]
		\centering
		\includegraphics[width=0.5\textwidth]{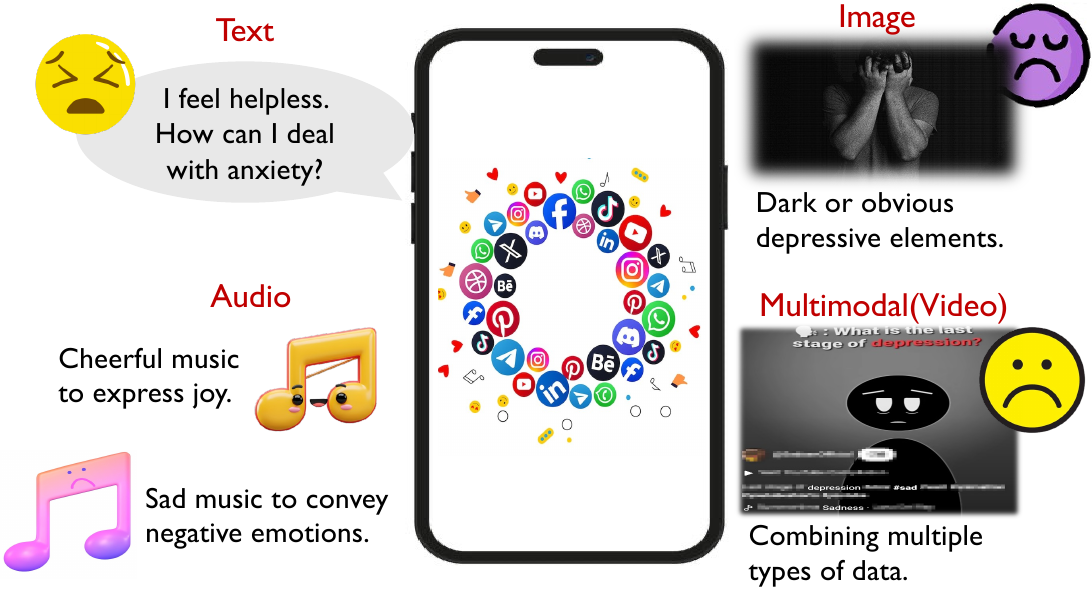}
		\caption{Representation of Different Data Types on Social Media.}
		\label{fig:fig3}
		\vspace{-10px}
	\end{figure}

	With the expansion of the Internet, a wide variety of social media contexts have emerged, each offering distinct modalities and user-generated content. 
	This diversity in data structures and contents provides a rich resource for mental disorder detection. 
	Figure~\ref{fig:fig3} illustrates how different types of data are represented on social media.
	The following is a detailed introduction to the types of data commonly found on mainstream platforms:
	\begin{itemize}[leftmargin=*] 
		\item \textbf{Text data}: On social media, text data typically appears in the form of tweets, status updates, comments, and more.
		\item \textbf{Audio data}: Audio data on social media primarily includes voice in videos, background music in shared content, etc.
		\item \textbf{Image data}: Image data on social media mainly includes selfies and user-posted pictures.
		\item \textbf{Multimodal data}: Multimodal data on social media combines text, voice, images, and other types of information.
	\end{itemize}

	\subsection{Large Language Models}

	\begin{figure}[t]
	\centering
	\includegraphics[width=0.5\textwidth]{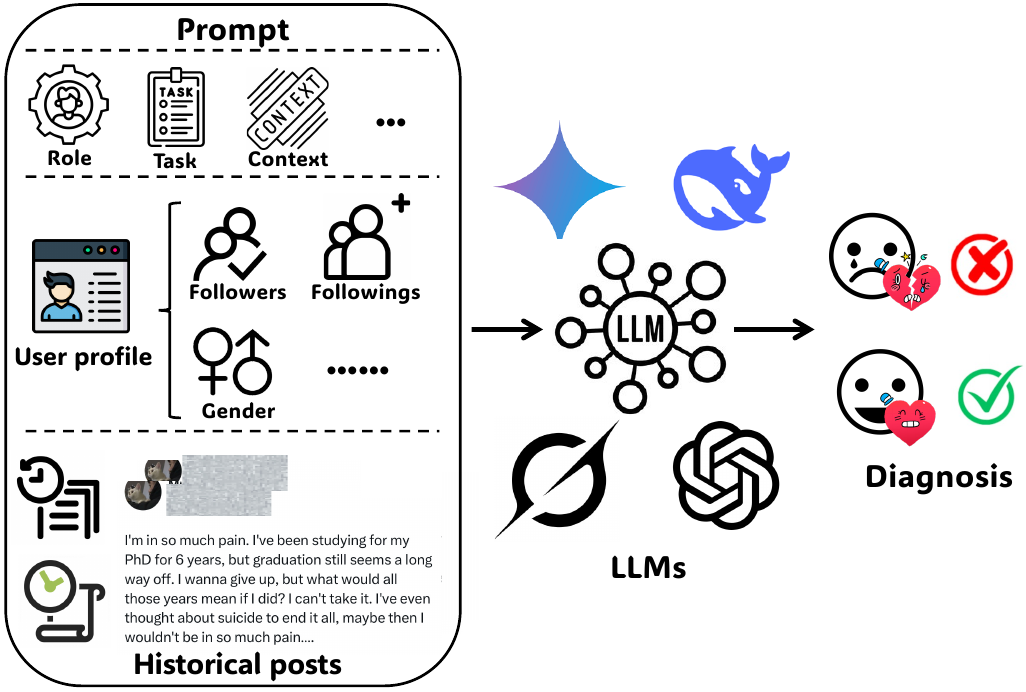}
	\caption{Overview of LLMs for Mental Disorder Detection.
	LLMs take a structured prompt that combines role, task, and contextual instructions with user profile information and historical posts, and then output a diagnosis.}
	\label{fig:fig4}
	\vspace{-5px}
	\end{figure}

	LLMs are trained on large-scale textual corpora to learn language patterns, syntax, semantics, and world knowledge, enabling natural language understanding, generation, and logical reasoning.
	Mainstream LLMs, such as the GPT series \cite{brown2020language,hurst2024gpt,achiam2023gpt}, LLaMA \cite{touvron2023llama,dubey2024llama}, DeepSeek \cite{liu2024deepseek,lu2024deepseek,guo2025deepseek,liu2024deepseekv3} and Qwen\cite{yang2025qwen3}, are based on the Transformer architecture~\cite{vaswani2017attention}, which follows a decoder-only structure that efficiently captures long-range dependencies in sequence data, significantly improving the computational efficiency and expressive power of the models.
	These LLMs have been successfully applied to a wide variety of tasks, including natural language processing \cite{min2023recent,xu2024large} and computer vision \cite{liu2024visual,zhang2024vision}.
	
	LLMs are highly effective for processing large-scale text data from diverse sources such as social media, blogs, and forums.
	They utilize self-attention mechanisms to capture long-distance dependencies and subtle nuances in language, enabling them to perform complex tasks like text categorization, sentiment analysis, and psychological assessments.
	Through pre-training, fine-tuning, and reinforcement learning from human feedback \cite{lee2023rlaif,ouyang2022training}, LLMs obtain impressive cross-task generalization, making them adaptable to a wide range of applications.
	Furthermore, their few-shot and zero-shot learning capabilities allow them to make inferences from minimal training examples\cite{brown2020language}, enhancing their flexibility for mental disorder prediction tasks.
	Some LLMs even extend their capabilities to multimodal data \cite{zhang2024mm,wu2023multimodal}, processing text, images, audio, and video simultaneously, thereby enriching their understanding and generation of information.
	Figure~\ref{fig:fig4} presents an overview of LLMs for mental disorder detection.

	\subsection{Retrieval-Augmented Generation}
\begin{figure}[t]
		\centering
		\includegraphics[width=0.48\textwidth]{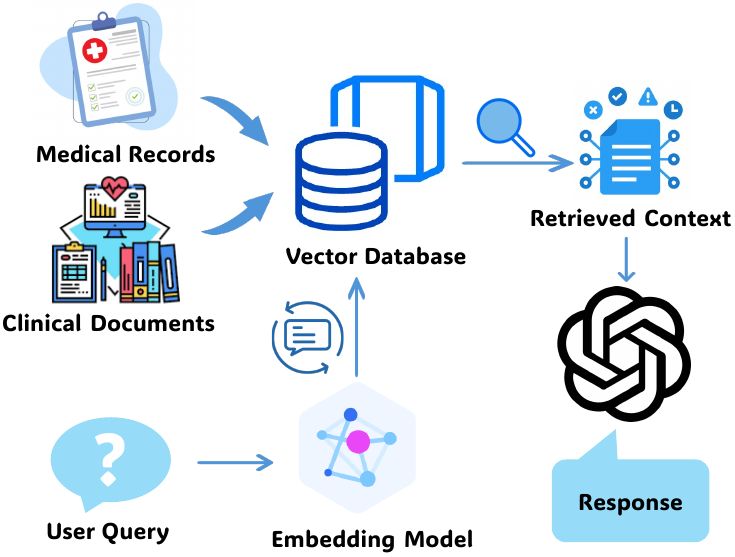}
		\caption{Illustration of the Basic Retrieval-Augmented LLMs framework. LLM retrieves external medical knowledge and then generates the final response.}
		\label{fig:fig5}
		\vspace{-8px}
	\end{figure}

	 Retrieval-Augmented Generation (RAG)\cite{lewis2020retrieval,gao2023retrieval} is an advanced artificial intelligence architecture designed to enhance the capabilities of LLMs by dynamically incorporating external knowledge bases.
	 Figure~\ref{fig:fig5} illustrates the basic workflow of RAG.
	 Its core workflow begins when a prompt is received, at which point the RAG system utilizes a retriever module to query and recall relevant contextual information from a large-scale, indexable knowledge source, such as specialized literature, databases, or web pages.
	 Subsequently, this retrieved information, along with the original prompt, is fed into a generator module (i.e., the LLM) to guide the model in producing a final, more information-rich response.

	 RAG technology offers several key advantages over standard LLMs\cite{fan2024survey,fan2025towards}. Firstly, it enhances factual accuracy and mitigates hallucinations\cite{wu2025multirag,shuster2021retrieval}.
	 By anchoring the LLM's generation process to specific, verifiable external facts, RAG significantly reduces the risk of the model fabricating incorrect information. Secondly, it enables dynamic knowledge updating.
	 Whereas the internal knowledge of an LLM is frozen post-training, RAG allows the system to access the most current information by simply updating the external knowledge base, thereby obviating the need for costly model retraining.
	 Finally, it improves transparency and interpretability\cite{ni2025towards}. 
	 Because responses are generated based on specific retrieved documents, the RAG system can cite its sources, which facilitates verification and traceability.

	 In the domain of mental health, these characteristics of RAG demonstrate significant application potential\cite{yang2025cascadercg,amugongo2025retrieval}. This field places exceptionally high demands on the reliability, accuracy, and evidence-based nature of information.
	 The core advantage of RAG is its capacity to ensure that LLM-generated responses are grounded in authoritative and professional knowledge sources, such as the criteria from the Diagnostic and Statistical Manual of Mental Disorders (DSM-5~\cite{apa2013diagnostic}), clinical practice guidelines, or recent peer-reviewed research.
	 This greatly mitigates the risk of the model providing erroneous medical advice or harmful information.
	 Furthermore, by accessing domain-specific knowledge bases, RAG can empower LLMs to deliver more in-depth psychoeducational content or assist in identifying complex mental health conditions, rendering them safer and more effective for tasks such as psychological support, preliminary screening, and resource navigation.

	\subsection{LLM-Based Agentic Systems}
\begin{figure}[t]
		\centering
		\includegraphics[width=0.49\textwidth]{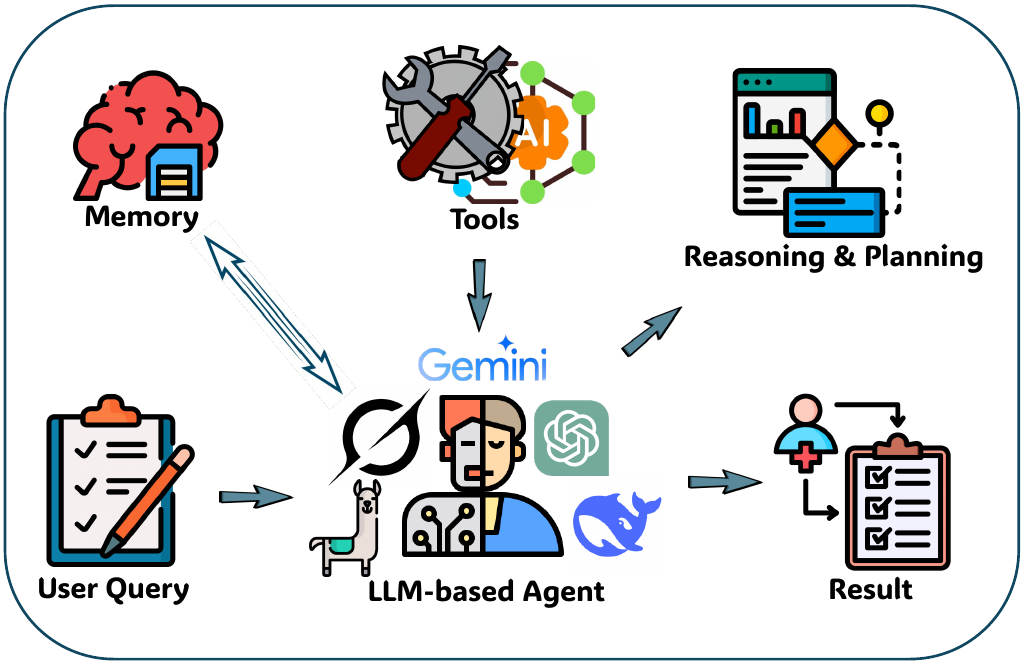}
		\caption{Illustration of the Basic LLM-based Agentic Systems.}
		\label{fig:fig6}
		\vspace{-10px}
	\end{figure}

	Building on the capabilities of LLMs and RAG technology, agentic systems represent a higher level of abstraction where the LLM is not just a generator but is embedded as the central reasoning and decision-making engine within a more extensive, goal-oriented system\cite{cheng2024exploring,liu2025advances}.
	An agentic system's primary distinction is its ability to operate autonomously and interactively to achieve a goal. 
	Figure~\ref{fig:fig6} shows the general framework of LLM-based agentic systems.
	These systems can maintain state and memory across extended interactions\cite{zhang2025survey}, allowing them to build context over time.
	Furthermore, agentic systems possess planning capabilities\cite{masterman2024landscape,huang2024understanding}, enabling them to decompose complex user goals or queries into a series of manageable subtasks.

	This creates a dynamic loop: the agentic system reasons about a goal, selects a tool, acts, observes the outcome, and then iteratively refines its next step based on that feedback.
	This ability to orchestrate complex, multi-step workflows and simulate decision-making makes agentic systems particularly powerful for navigating the dynamic and highly personalized nature of mental health challenges. 
	Applied to mental disorder detection on social media, these properties enable agentic systems to move beyond single-post classification toward more robust, context-aware monitoring pipelines that better capture temporal patterns, reduce false positives and false negatives, and incorporate external knowledge in a controllable way.

		\section{Methodologies for Mental Disorder Detection across Clinical Categories} \label{sec:3}
	This section introduces the applications of LLMs in three categories of mental disorders (Sec.~\ref{subsec:disor}).
	For areas where relevant research is lacking, we propose potential future research directions, emphasizing their strengths and values.
	Figure \ref{fig:fig7} summarizes the research on these three categories of mental disorders based on different methodologies.

	\subsection{Problem Definition}
\begin{figure}[htbp]
		\centering
		\includegraphics[width=0.5\textwidth]{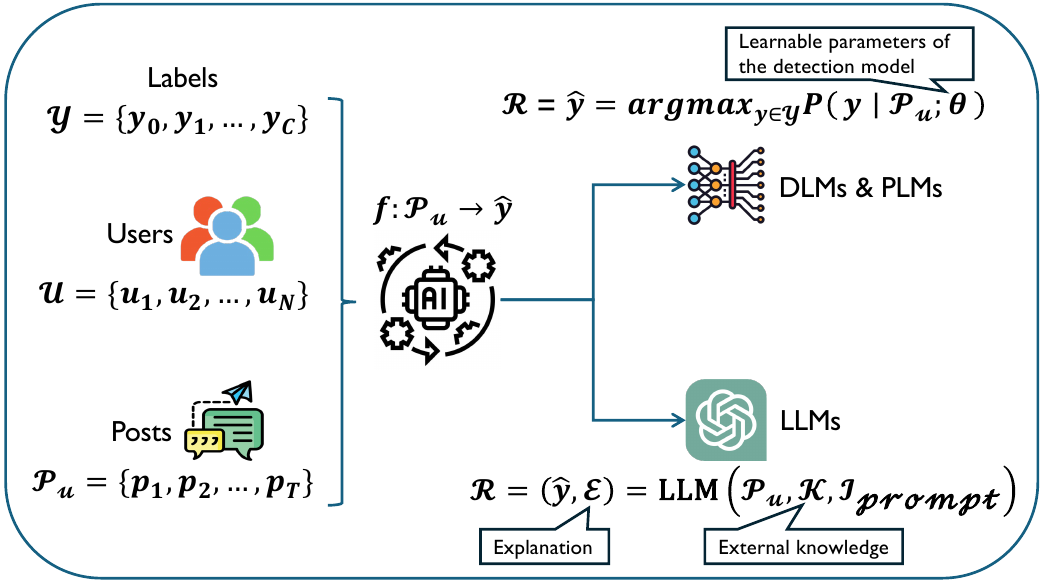}
		\caption{Overall Framework for Detecting Mental Disorders on Social Media. This schema illustrates the process of mapping user posts to diagnostic labels using either learnable detection models or LLMs.}
		\vspace{-6px}
		\label{fig:fig8}
\end{figure}

	Figure~\ref{fig:fig8} illustrates a framework for detecting mental disorders based on social media.
	Let $\mathcal{U} = \{u_1, u_2, \dots, u_N\}$ denote a set of social media users. 
	For each user $u \in \mathcal{U}$, we observe a chronological sequence of historical posts $\mathcal{P}_u = \{p_1, p_2, \dots, p_T\}$, where $T$ represents the number of posts. 
	Let $\mathcal{Y} = \{y_0, y_1, \dots, y_C\}$ be the set of predefined severity levels for a specific mental disorder (e.g., mild, moderate, or severe).

	Given the historical posts sequence $\mathcal{P}_u$ of user $u$, the goal is to learn a mapping function $f: \mathcal{P}_u \rightarrow \hat{y}$, which predicts the most probable mental state $\hat{y} \in \mathcal{Y}$ for the user. 

	\textbf{Deep learning and pre-trained language models.}
	The model typically encodes the user's historical posts sequence $\mathcal{P}_u$ into a dense vector and maps it to a probability distribution over the predefined categories $\mathcal{Y}$.
	Formally, the objective is to maximize the conditional probability:
	\begin{equation}
		\hat{y} = \operatorname*{argmax}_{y \in \mathcal{Y}} P(y \mid \mathcal{P}_u; \theta),
	\end{equation}
	where $\theta$ represents the learnable parameters of the detection model (e.g., weights in the transformer layers).

	\textbf{LLMs.}
	Given the user's posts $\mathcal{P}_u$ and potentially retrieved external knowledge $\mathcal{K}$ (e.g., DSM-5 criteria via RAG), the model is tasked with generating a response $\mathcal{R} = (\hat{y}, \mathcal{E})$, where $\mathcal{E}$ is a natural language explanation grounding the diagnosis.
	The generation process can be formulated as:
	\begin{equation}
		(\hat{y}, \mathcal{E}) = \operatorname{LLM}(\mathcal{P}_u, \mathcal{K}, \mathcal{I}_{prompt}),
	\end{equation}
	where $\mathcal{I}_{prompt}$ denotes the instructional prompt guiding the LLM to perform reasoning.

\begin{figure*}[htbp]
    \centering
    \resizebox{0.95\textwidth}{!}{

		\begin{forest}
		for tree={
			grow=east,
			parent anchor=east,
			child anchor=west,
			draw=black,
			rounded corners=3pt,
			line width=0.6pt,
			align=left,
			edge={black, line width=0.6pt},
			l sep=1cm,
			s sep=0.4cm,
			drop shadow,
			tier/.option=level,
			where level=0{
				anchor=center
			}{
				anchor=west
			},
		},
		where level=0{
			fill=rootgreen, 
			text width=1.8cm, 
			font=\rmfamily\bfseries\footnotesize\linespread{0.6}\selectfont,
		}{},
		where level=1{
			fill=catpink, 
			text width=1.6cm,
			font=\rmfamily\bfseries\footnotesize\linespread{0.6}\selectfont,
		}{},
		where level=2{
			fill=suborange, 
			font=\rmfamily\footnotesize\linespread{0.6}\selectfont,
		}{},
		where level=3{
			fill=leafblue,
			draw=black!60,
			font=\footnotesize\rmfamily\linespread{0.6}\selectfont,
			l sep=0.8cm,
			s sep=0.3cm,
		}{}
		[Methodologies
			[Externalizing\\Disorders
				[LLM-Based Agentic Systems
					[{Lin et al.\cite{lin2025ask}}]
				]
				[RAG-Driven LLMs
					[{Yao et al.\cite{yao2024personalised}}]
				]
				[Large Language Models
					[{Zhu et al.\cite{zhu2025leveraging}, Vanpech et al.\cite{vanpech2024detecting},\\
					García-Méndez et al.\cite{garcia2024promoting}, Jiang\cite{jiang2025social}}]
				]
				[DLMs \& PLMs
					[{Guntuku et al.\cite{guntuku2019language}, Chatzakou et al.\cite{chatzakou2019detecting}, Alsharif et al.\cite{alsharif2024adhd},\\
					Mali et al.\cite{mali2025automatic}, Perera et al.\cite{perera2024cyberbullying}, López-Vizcaíno et al.\cite{lopez2021early},\\
					Prabhu et al.\cite{prabhu2025comprehensive}, Bozyiğit et al.\cite{bozyiugit2021cyberbullying}, Pericherla et al.\cite{pericherla2024transformer},\\
					Lee et al.\cite{lee2024detecting}, Sihab-Us-Sakib et al.\cite{sihab2024cyberbullying}, Chen et al.\cite{chen2023exploring},\\
					Murshed et al.\cite{murshed2023faeo}}]
				]
			]
			[Psychotic\\Disorders
				[Large Language Models
					[{Liu et al.\cite{jiaying2025displaying}}]
				]
				[DLMs \& PLMs
					[{Birnbaum et al.\cite{birnbaum2017collaborative}, Plank et al.\cite{plank2024reduced}, Kim et al.\cite{kim2020deep},\\
					Bae et al.\cite{bae2021schizophrenia}, McManus et al.\cite{mcmanus2015mining}, Jeong et al.\cite{jeong2023exploring},\\
					Mitchell et al.\cite{mitchell2015quantifying}}]
				]
			]
			[Internalizing\\Disorders
				[LLM-Based Agentic Systems
					[{Liu et al.\cite{liu2025pychoagent}, Shafi\cite{shafi2025wellbeingagent}}]
				]
				[RAG-Driven LLMs
					[{Xu et al.~\cite{xu2024utilizing}, Nushida et al.~\cite{nushida2025automated}, Antony et al.~\cite{antony2025retrieval}, \\
					Ravenda et al.~\cite{ravenda2025llms}, Kermani et al.~\cite{kermani2025systematic}, Wang et al.\cite{wang2025posts}}]
				]
				[Large Language Models
					[{Sabaneh et al.~\cite{sabaneh2023early}, Wang et al.~\cite{wang2024explainable}, Shin et al.~\cite{shin2024using}, \\
					Singh et al.~\cite{singh2024extraction}, Qin et al.~\cite{qin2023read}, Lamichhane~\cite{lamichhane2023evaluation}, \\
					Bhaumik et al.~\cite{bhaumik2023mindwatch}, Qi et al.~\cite{qi2023supervised}, Song et al.~\cite{song2024combining}, \\
					Alhamed et al.~\cite{alhamed2024classifying}, Lan et al.~\cite{lan2024depression}, Liu et al.~\cite{liu2024enhancing}, \\
					Bauer et al.~\cite{bauer2024using}, Lashgari et al.\cite{lashgari2025sentinel}, Xu et al.~\cite{xu2024mental}, \\
					Shah et al.\cite{shah2025advancing}, Zhu et al.\cite{zhu2025social}, Zheng et al.\cite{zheng2024cascade}, \\
					Radwan et al.\cite{radwan2024predictive}, Qi et al.\cite{qi2025supervised}, Nguyen et al.\cite{nguyen2025supporters}, \\
					Liu et al.~\cite{liu2024emollms}}]
				]
				[Pre-trained Language Models
					[{Wang et al.~\cite{wang2020depression}, Metzler et al.~\cite{metzler2022detecting}, Owen et al.~\cite{owen2023enabling},\\
					Verma et al.~\cite{verma2023ai}, Wu et al.\cite{wu2025psychological}, Leow et al.\cite{leow2025comparison},\\
					Lee et al.~\cite{lee2023towards}, Wang et al.~\cite{wang2023contrastive}, Ragheb et al.~\cite{ragheb2021negatively}}]
				]
			]
		]
		\end{forest}
	}
	\caption{Taxonomy of Research Methodologies under Different Mental Disorders.}
    \label{fig:fig7}
\end{figure*}

	\subsection{Methodologies for Internalizing Disorders}
	In terms of internalizing disorders, anxiety and depression often co-occur, and the depressive episode of bipolar disorder exhibits similar manifestations to depression.
	Additionally, individuals with depression face a higher risk of suicide, with some studies indicating that about 60\% of people who commit suicide suffer from depression~\cite{law2008suicide}. Therefore, beyond specific mental disorders, suicide prediction itself has also become a prominent research task.

\subsubsection{\textbf{Pre-trained Models}}
Early efforts in utilizing pre-trained models for internalizing disorders focused on basic detection tasks using BERT variants on social media text. 
Wang et al.~\cite{wang2020depression} pioneered this by applying BERT, RoBERTa, and XLNet to predict depression levels in 13,993 microblogs, establishing a baseline for automated screening.
To address the challenges of model subjectivity and generalization, Ragheb et al.~\cite{ragheb2021negatively} proposed the negatively correlated noisy learners architecture, which ensembles noisy base learners on top of transformer-based backbones (BERT, RoBERTa, XLNet) to enhance diversity for detecting depression, anorexia, and self-harm. 
Simultaneously, to improve semantic representations in sparse data scenarios, Wang et al.~\cite{wang2023contrastive} enhanced semantic representations in sparse data via a self-supervised contrastive learning framework, employing auxiliary discrimination tasks to explicitly capture nuances of stressors and stressful emotions.
Building on this foundation, Metzler et al.~\cite{metzler2022detecting} refined suicide-related content detection with BERT and XLNet, introducing diversified labeling for nuanced categories to overcome binary limitations.
Subsequently, Owen et al.~\cite{owen2023enabling} incorporated temporal dynamics using BERT, ALBERT, and domain-adapted variants like MentalBERT\cite{ji2021mentalbert}, enabling early depression identification through language change analysis, which addressed static post evaluations.
Further advancing temporal modeling, Lee et al.~\cite{lee2023towards} utilized Sentence-BERT within a multi-task learning framework that features a temporal symptom-aware attention mechanism to dynamically weigh historical posts for predicting future suicidality in bipolar disorder patients.
Moreover, Verma et al.~\cite{verma2023ai} enhanced interpretability with RoBERTa on 27,972 mental health entries and 7,650 Reddit posts, analyzing linguistic profiles to provide deeper cognitive insights beyond mere classification.
To enhance detection capabilities in social networks, Wu et al.~\cite{wu2025psychological} proposed a multi-level transfer learning framework that integrates BERT with mental health knowledge embeddings and hierarchical graph convolution, effectively identifying subtle psychological crisis signals within noisy social network data.

\subsubsection{\textbf{Large Language Models}}
As research progressed, studies shifted towards leveraging LLMs like GPT for more flexible and context-aware tasks, overcoming the rigidity of pre-trained models.
Sabaneh et al.~\cite{sabaneh2023early} leveraged GPT-3.5\cite{brown2020language} for depression detection in 1,058 Arabic tweets, optimizing with UMLS\cite{bodenreider2004unified} and vector techniques to handle low-resource languages effectively.
To further improve explainability, Wang et al.~\cite{wang2024explainable} employed LLaMA-2 and variants on 3,107 TREC files and 170 users' writings, correlating posts with BDI\cite{steer1999common} scales via LLM evaluations, mitigating the lack of clinical grounding in prior approaches.
Similarly, Shin et al.~\cite{shin2024using} utilized GPT-3.5 and GPT-4 on 91 participants' 428 diaries, scoring depression and suicide risk with PHQ-9 and BSS for quantifiable assessments, building on detection by integrating standardized tools.
Additionally, Singh et al.~\cite{singh2024extraction} applied Mixtral\cite{jiang2024mixtral} and Tulu-2-DPO-70B\cite{ivison2023camels} with prompting strategies on 934 users' data, extracting suicidal ideation evidence to enhance evidential summaries over basic risk scoring.
Addressing scalability issues, Radwan et al.~\cite{radwan2024predictive} utilized GPT-3 embeddings as feature inputs for traditional machine learning classifiers such as SVM to detect stress disorders from Reddit posts, demonstrating a predictive analytics approach that effectively handles large-scale social media data.
To further refine detection through specialized features, Zhu and Huang~\cite{zhu2025social} introduced a framework that fuses deep semantic embeddings from LLaMA-2 with psychological features extracted via lexicons like LIWC, aiming to capture latent psychological distortions in social media texts that purely semantic models might miss.

To leverage the strengths of both paradigms, hybrid methods emerged, combining the efficiency of pre-trained models with the generative prowess of LLMs, addressing the limitations of using either in isolation.
Qin et al.~\cite{qin2023read} integrated ChatGPT-3.5, GPT-3, and BERT for multimodal explainable depression detection on 2,000 users' text-image data, using DSM-5\cite{apa2013diagnostic} criteria and chain-of-thought prompting to add reasoning depth.
Following this, Lamichhane~\cite{lamichhane2023evaluation} merged GPT-3.5's zero-shot classification with BERT on 3,553 posts for multi-disorder tasks, improving adaptability in data-scarce scenarios.
To scale for larger datasets, Bhaumik et al.~\cite{bhaumik2023mindwatch} fine-tuned ALBERT and Bio-Clinical BERT\cite{alsentzer2019publicly} alongside GPT-3.5 prompting on 232,000 posts for suicide risk, hinting at retrieval enhancements.
Qi et al.~\cite{qi2023supervised} advanced this with diverse prompting on BERT, GPT variants, and LLaMA-2 for cognitive distortions detection in 1,249 and 3,407 posts, refining hybrid strategies for multi-label accuracy. 
Expanding on comparative analyses, Leow et al.~\cite{leow2025comparison} contrasted fine-tuned BERT with zero-shot BART\cite{lewis2020bart} on a depression dataset, highlighting that domain-specific fine-tuning remains crucial for capturing subtle depression indicators in social media, outperforming zero-shot generative approaches.
Providing a comprehensive benchmark for Chinese social media, Qi et al.~\cite{qi2025supervised} compared supervised learning with various LLMs under zero-shot, few-shot, and fine-tuning settings, revealing that although fine-tuning substantially improves LLM performance, supervised models like BERT still outperform them on complex cognitive distortion detection tasks.

Further extending these hybrid strategies, Song et al.~\cite{song2024combining} fused BART-based VAE with LLaMA-2 for timeline summarization on 500 Talklife users, addressing longitudinal gaps in snapshot hybrids.
Extending to user-level changes, Alhamed et al.~\cite{alhamed2024classifying} combined Alpaca, BERT variants, and GPT-3.5 on 120 users' 1.9 million tweets, detecting pre- and post-diagnosis shifts for more reliable predictions.
Lan et al.~\cite{lan2024depression} further incorporated medical knowledge with GPT-3.5 and MentalRoBERTa on 2,000 users' 1.38 million posts, enabling mood course modeling and explanations.
In parallel, Liu et al.~\cite{liu2024enhancing} proposed multi-task learning with BERT variants, GPT-4, and MentaLLaMA\cite{yang2024mentallama} on 54,412 and 8,554 posts, tackling simultaneous classifications to mitigate siloed tasks.
Bauer et al.~\cite{bauer2024using} utilized BERT embeddings with GPT-4 and ProtoDash\cite{gurumoorthy2019efficient} on 2.9 million posts for suicidality prototypes, providing extreme case insights.
Enhancing the capabilities of few-shot learning, Nguyen et al.~\cite{nguyen2025supporters} employed few-shot in-context learning with models like GPT-4 to detect mental health misinformation on video-sharing platforms, demonstrating the effectiveness of ICL in analyzing stigmatizing language in comments where labeled data is scarce.
Similarly, Zheng et al.~\cite{zheng2024cascade} designed a cascade model that uses GPT-3.5 via in-context learning to extract behavioral feature scores, which are then processed by a lightweight neural network, effectively reducing systematic errors and improving depression detection accuracy on Weibo data.

Finally, Xu et al.~\cite{xu2024mental} unified zero/few-shot prompting across BERT, FLAN-T5, and GPT variants on seven datasets, demonstrating that fine-tuning LLMs far outperforms prompt engineering.
Confirming the benefits of domain adaptation, Shah et al.~\cite{shah2025advancing} showed that parameter-efficient fine-tuning of GPT-3.5 Turbo and LLaMA2-7B significantly boosts depression detection accuracy on Twitter compared to general base models.
To address the limitations of existing models in handling fine-grained affective regression alongside classification, Liu et al.~\cite{liu2024emollms} introduced EmoLLMs, a series of instruction-tuned open-source models (e.g., LLaMA2, OPT) developed via a multi-task instruction dataset, achieving generalization capabilities comparable to GPT-4 in complex emotional analysis.
Similarly, Lashgari et al.~\cite{lashgari2025sentinel} introduced the SENTINEL-LLM framework, which ensembles multiple fine-tuned LLMs (e.g., Qwen2.5, LLaMA3) and incorporates a weighted “frequent-rare" dictionary to enhance sensitivity to suicidal ideation patterns in social media text.

These studies illustrate the evolving role of LLMs and deep learning in mental health research.
While pre-trained models, such as BERT and MentaLLaMA, continue to demonstrate high accuracy in depression detection, direct classification using LLMs remains unreliable due to inconsistencies in cue-based approaches.
LLMs like GPT have been fine-tuned with natural conversation as one of the optimization goals, but have not yet demonstrated strong performance for specific tasks such as classification prediction.
Most research has utilized these LLMs to perform auxiliary tasks, such as annotation and data enhancement, while reserving specific task execution for other models.
Moreover, LLMs like GPT also contribute by providing explanations for predictions, which enhances the interpretability of results, making the process more transparent and clinically meaningful. 

\subsubsection{\textbf{RAG-Driven LLMs}}
This section explores the integration of RAG with LLMs to enhance mental health applications, primarily focusing on detection, diagnosis, screening, and therapeutic support through analysis of social media text, questionnaires, and knowledge bases, with common themes including leveraging RAG to improve LLM performance by incorporating external knowledge like knowledge graphs, retrieved examples, or psychological rules, thereby addressing limitations such as hallucinations, lack of personalization, and interpretability in mental health contexts.

To reduce hallucinations, Xu et al.~\cite{xu2024utilizing} fine-tuned an LLM with RAG on Kaggle comments for suicide detection, retrieving suicidal language traits from Reddit knowledge bases to ground predictions.
To structure knowledge better, Nushida et al.~\cite{nushida2025automated} employed GPT with GraphRAG on social media graphs for depression detection, contrasting depressive and healthy data for enhanced accuracy.
Building on this, Antony and Schoene~\cite{antony2025retrieval} applied LLaMA\cite{dubey2024llama} with vector search on CLPsych timelines for self-state assessment, using in-context examples to enable dynamic pattern recognition over static retrieval.
Similarly, Ravenda et al.~\cite{ravenda2025llms} used GPT-4 with adaptive RAG on social content for multi-disorder screening, predicting questionnaire responses like BDI-II to broaden clinical applicability.
Moreover, Kermani et al.~\cite{kermani2025systematic} leveraged LLaMA-3 with clinical document retrieval on 20,000 tweets and 54,412 posts, refining emotion and condition classifications through domain-specific grounding.
Addressing the need for longitudinal analysis, Wang et al.~\cite{wang2025posts} proposed a hybrid LLM method that uses RAG to retrieve relevant historical posts from user timelines, assisting the model in generating longitudinal summaries of mental health dynamics and identifying adaptive evidence.

Overall, these studies highlight RAG's potential in bridging computational efficiency with clinical relevance by augmenting LLMs with domain-specific knowledge to support detection, screening, and therapeutic applications.
However, significant limitations persist, including small or inconsistent sample sizes that hinder generalizability, a heavy reliance on social media data that is prone to biases and noise, and an insufficient evaluation of long-term real-world efficacy in clinical settings.

\subsubsection{\textbf{LLM-based Agentic Systems}}
	Recent research highlights the potential of this approach for dynamic and personalized mental health support.
	Liu et al.~\cite{liu2025pychoagent} proposed PsychoAgent, a psychology-driven framework that employs role-playing agents to simulate users' mental processes during disasters. 
	By integrating standardized risk scales~\cite {mclennan2020conceptualising}, the system moves beyond reactive dialogue to provide interpretable panic predictions.
	Similarly, Shafi~\cite{shafi2025wellbeingagent} introduced WellbeingAgent, an architecture where an orchestrator dynamically determines whether to engage in conversation or invoke external functions to retrieve real-time physiological data, thereby grounding responses in user-specific context to reduce hallucinations.
	Unlike single-turn interactions, these agentic systems maintain state over time and coordinate specialized components, such as risk assessment or emotion regulation modules, to orchestrate complex workflows resembling multidisciplinary clinical care.

	This allows the LLM's responses to be grounded, incorporating the user's historical data, thus achieving truly personalized support and reducing hallucinations.
	Crucially, effective agentic systems must incorporate robust safety layers, including explicit crisis-detection routines, escalation pathways to human professionals, and conservative policies that avoid making definitive diagnoses or therapeutic decisions beyond their scope.
	When combined with careful human oversight, privacy-preserving data management, and culturally sensitive design, agentic systems have the potential to transform LLMs and RAG from static information providers into dynamic, context-aware collaborators that augment clinical care and broaden access to timely, personalized mental health support.

\subsection{Methodologies for Psychotic Disorders}
	Psychotic disorders are serious mental disorders characterized by significant impairments in thinking, perception, and emotions.
    The social burden imposed by psychotic disorders is substantial \cite{rossler2005size}. Society must not only bear the direct costs of treatment and long-term care, but also indirect burdens such as productivity losses due to disability and premature death, increased strain on caregiving and nursing staff, and profound impacts on families and communities.

	Schizophrenia is widely recognized as the most primary and representative condition within the category of psychotic disorders \cite{mccutcheon2020schizophrenia}. 
    Furthermore, delusional disorder is classified under the ``Schizophrenia Spectrum and Other Psychotic Disorders'' in diagnostic frameworks such as the DSM-5\cite{apa2013diagnostic}, as it shares the core positive symptom of delusions. 
    Consequently, existing computational research predominantly targets schizophrenia as the prototypical disorder for detecting psychosis. 
	Therefore, the following studies reviewed in this section primarily focus on schizophrenia detection.

	\subsubsection{\textbf{DLMs \& PLMs}}
	Early research on schizo\-phrenia detection primarily relied on manual feature extraction combined with traditional machine learning algorithms, gradually evolving toward deep learning and pre-trained models to capture more complex semantic features. 
	In the phase of feature engineering, Mitchell et al.\cite{mitchell2015quantifying} employed LDA and brown clustering alongside SVM to analyze Twitter data, aiming to overcome data scarcity and the unstructured nature of social media text by identifying specific linguistic patterns such as irrealis mood and lack of emoticons. 
	Similarly, McManus et al.\cite{mcmanus2015mining} utilized SVM and neural networks on Twitter data to address the challenge of identifying at-risk individuals, relying on features such as emoticon usage, time-of-day posting patterns, and dictionary terms to distinguish users with schizophrenia. 
	To further address the issue of noise in online self-disclosure data, Birnbaum et al.\cite{birnbaum2017collaborative} adopted a collaborative approach that combined RF with expert clinical appraisals on Twitter data, successfully improving classification accuracy by identifying differences in personal pronoun usage and perceptual word categories. 
	Subsequently, Bae et al.\cite{bae2021schizophrenia} applied random forests and unsupervised clustering techniques (t-SNE and DBSCAN) to Reddit posts, aiming to fill the gap in understanding specific content topics (e.g., hallucinations, medication); they identified increased usage of third-person plural pronouns and negative emotion words as key detection signals.

	While these methods are effective, they have limitations. 
	For instance, the performance of machine learning models like SVM and ANN often depends on the quality of feature engineering and may fail to capture deeper linguistic dependencies and sentiment changes.

	As deep learning technologies advanced, research shifted from manual feature engineering to automatic feature learning. Kim et al.\cite{kim2020deep} proposed a model based on CNNs combined with Word2Vec embeddings to process Reddit data. 
	This approach addressed the difficulty of noisy data caused by comorbidities by learning the specific language styles of mental health subreddits, enabling the accurate classification of multiple disorders, including schizophrenia. 
	In the realm of PLMs, the focus turned to quantifying text coherence and deep semantics. Jeong et al.\cite{jeong2023exploring} utilized the next-sentence prediction and surprisal capabilities of BERT to analyze speech transcripts from clinical interviews, aiming to mitigate the subjectivity of traditional clinical rating scales by objectively quantifying specific symptoms, such as derailment and alogia. 
	Furthermore, Plank and Zlomuzica\cite{plank2024reduced} extended this coherence analysis to large-scale social media data, using the universal sentence encoder\cite{cer2018universal} to embed Reddit posts and calculate semantic similarity, revealing significantly reduced speech coherence in users with schizophrenia and dissociative identity disorder.

	\subsubsection{\textbf{Large Language Models}}
	Recent studies have begun to explore the application of LLMs in understanding multimodal data and complex emotional narratives. 
	Liu et al.\cite{jiaying2025displaying} introduced a multimodal LLM framework that combines LLaVA-1.6\cite{liu2023visual} for visual keyframe analysis with GPT-4o-mini\cite{hurst2024gpt} for textual processing. 
	Using YouTube vlogs created by individuals with schizophrenia, the study examined how fear, sadness, and joy are conveyed through visual composition and verbal narration, and found that visual appeal strongly influences audience engagement.
	
	
	
	However, the use of LLMs for detecting schizophrenia on social media is still in its nascent stages. 
	Despite this, LLMs demonstrate significant potential in handling the complexity of natural language, identifying subtle emotional shifts, and enabling automated analysis. 
	Notably, advanced paradigms such as RAG and agentic systems remain largely unexplored in this domain. 
	The integration of RAG could mitigate hallucinations by grounding generation in clinical guidelines, while agentic systems could enable more dynamic, multi-step diagnostic reasoning. With further domain adaptation, coupled with these emerging technologies, LLMs are expected to evolve into a powerful, reliable tool for detecting schizophrenia and other mental health issues in the future.
	

	\subsection{Methodologies for Externalizing Disorders}
	Beyond the specific scope of ADHD, it is crucial to contextualize online aggression within the broader spectrum of externalizing disorders.
	For instance, adolescents with ADHD exhibit significantly elevated rates of cyberbullying perpetration compared with typically developing peers, especially when comorbid externalizing problems, problematic Internet use, and moral disengagement are present~\cite{Yen2014,Liu2021,Pineda2024}. 
	In addition, ADHD users are less agreeable, more open, post more frequently, and use more negative words, fewer mild words, and profanity\cite{guntuku2019language}.
	Youth with conduct disorder, a prototypical externalizing condition characterized by persistent violation of social norms and others’ rights, also show markedly higher involvement in cyberbullying as perpetrators (and often as bully-victims) relative to controls~\cite{SchultzeKrumbholz2023,Healthspring_CD}. 
	More broadly, large-scale survey data indicate that higher levels of externalizing problems are robustly associated with increased engagement in bullying perpetration~\cite{LucasMolina2021}. 

	\subsubsection{\textbf{DLMs \& PLMs}}
	Researchers have continuously explored how to utilize deep learning and pre-trained models to more accurately capture temporal, linguistic, behavioral, and latent semantic features to address specific challenges in social media analysis. 
	For the detection of adult ADHD, Guntuku et al.\cite{guntuku2019language} utilized LIWC dictionaries and LDA topic modeling to extract linguistic features, revealing that patients exhibit more emotional dysregulation and self-criticism on social media, distinguishing them primarily through language patterns and posting behaviors. 
	To address the limitations of relying solely on text, Bozyiğit et al.\cite{bozyiugit2021cyberbullying} focused on cyberbullying detection by incorporating social media metadata alongside text mining, experimentally proving that these non-textual features significantly enhance machine learning classifier performance. 
	Addressing the evolution of language over time, Perera and Fernando\cite{perera2024cyberbullying} proposed a system based on supervised learning that combines TF-IDF, N-grams, and profanity detection to resolve confusion between extreme vocabulary and sarcastic contexts in cyberbullying.

	To further tackle the challenges of noise and implicit expressions in social media, Murshed et al.\cite{murshed2023faeo} proposed a hybrid model named FAEO-ECNN, which combines fuzzy adaptive equilibrium optimization for latent topic discovery with an extended CNN using wavelet pooling and rain optimization for precise detection of cyberbullying categories.

	With the advancement of deep learning, sequence models and Transformer architectures have been introduced to capture temporal dependencies and deep semantics. 
	To address the lack of semantic understanding in cyberbullying detection, Mali et al.\cite{mali2025automatic} proposed a hybrid model using BERT as a base classifier for aggressive behavior, combined with stacked bidirectional GRU and attention mechanisms to learn sequential semantics and spatial location information.
	Alsharif et al.\cite{alsharif2024adhd} compared machine learning and deep learning models on a Reddit dataset, utilizing TF-IDF for text feature extraction to diagnose ADHD, finding that while DL models were effective, RF performed best on their specific dataset. 
	Furthermore, focusing on the prediction of implicit symptoms, Lee et al.\cite{lee2024detecting} fine-tuned RoBERTa on Reddit posting histories to predict future comorbid ADHD diagnoses from users’ current descriptions of anxiety symptoms.

	In multimodal and low-resource language scenarios, Prabhu and Seethalakshmi\cite{prabhu2025comprehensive} proposed the MHSDF framework, innovatively fusing CNNs for spatial feature extraction and LSTMs for temporal dependency modeling, relying on the integration of text, image, and audio cues to detect complex hate speech involving sarcasm and metaphors. 
	Addressing detection bottlenecks in low-resource languages, Sihab-Us-Sakib et al.\cite{sihab2024cyberbullying} developed a Bengali cyberbullying dataset and compared various ML and DL models, finding that XLM-RoBERTa performed best in capturing complex grammar and context. 
	To optimize feature representation, Pericherla and Ilavarasan\cite{pericherla2024transformer} addressed the context-independent nature of traditional word embeddings by using RoBERTa to generate dynamic embeddings combined with LightGBM. 
	Additionally, regarding the behavioral analysis of ADHD patients on social media, Chen et al.\cite{chen2023exploring} employed DistilBERT\cite{sanh2019distilbert} for fine-grained sentiment analysis alongside Top2Vec topic modeling to analyze tweet content and metadata. 
	For the timeliness of cyberbullying detection, López-Vizcaíno et al.\cite{lopez2021early} proposed threshold and dual models utilizing bag-of-words and time-interval features, achieving early detection of cyberbullying incidents on the Vine social network. 
	To handle complex role classification in cyberbullying (e.g., bullies vs. aggressors), Chatzakou et al.\cite{chatzakou2019detecting} combined user, text, and network features, using RNNs and Word2Vec embeddings to capture semantic context for distinguishing between normal and abusive users.

	\subsubsection{\textbf{Large Language Models}}
	With the rise of LLMs, research focus has shifted from simple classification to leveraging the reasoning, generative, and multimodal capabilities of LLMs to address complex and dynamic detection tasks. 
	Addressing the challenge of symptom heterogeneity in ADHD diagnosis, Zhu et al.\cite{zhu2025leveraging} proposed an ensemble framework combining zero-shot reasoning from LLaMA3\cite{dubey2024llama}, fine-tuned RoBERTa, and traditional SVM, achieving robust classification using narrative text data. 
	To balance content semantics and network structure in large-scale social networks, Jiang and Ferrara\cite{jiang2025social} proposed Social-LLM, which fine-tunes SBERT-MPNet combined with network homophily for contrastive learning, enabling scalable user behavior detection (e.g., political leaning, toxicity) without relying on full-graph training.

	Targeting semantic understanding in multimodal content, Vanpech et al.\cite{vanpech2024detecting} utilized GPT-4 Vision\cite{lyu2025gpt} to generate detailed descriptive metadata for images, combined with a custom LLM for classification, effectively solving the identification of potential cyberbullying content in Twitter images. 
	To adapt to the rapid evolution of social media content and provide interpretability, García-Méndez and De Arriba-Pérez\cite{garcia2024promoting} proposed a streaming machine learning framework that uses GPT-4o-mini and NLP techniques for feature engineering to extract high-level reasoning features and generate natural language explanations.

	\subsubsection{\textbf{RAG-Driven LLMs}}
	To handle personalized and subjective tasks where general-purpose models face limitations, RAG has been introduced to integrate external knowledge.
	Addressing the issue of subjective bias in abusive language detection caused by individual personality differences, Yao et al.\cite{yao2024personalised} proposed a RAG-based solution that mines association rules from psychological datasets and retrieves them as external knowledge, guiding GPT-3.5 Turbo to make personalized abusive language judgments based on user psychological traits.

	\subsubsection{\textbf{LLM-based Agentic Systems}}
	For complex tasks involving highly implicit or multimodal content, LLM-based agentic systems have demonstrated understanding capabilities surpassing single models. 
	Targeting implicit social abuse content in image-text memes, Lin et al.\cite{lin2025ask} designed the ``Ask, Acquire, Understand" multi-agent framework, where LLMs act as agents with distinct roles that actively question a vision expert model (e.g., LLaVA\cite{liu2023visual}), utilizing multi-turn conversational reasoning to continuously acquire deep visual information, thereby significantly enhancing the detection of potential harm in memes under zero-shot settings.

	However, despite these pioneering efforts, the application of RAG and LLM-based agentic systems in the domain of cyberbullying and social abuse detection remains largely underexplored.
	Current literature predominantly focuses on standard fine-tuning or prompting paradigms, with scant research investigating how external knowledge retrieval or collaborative agent interactions can address the rapid evolution and context-dependence of online abuse. 
	This scarcity represents a significant gap, as static models often struggle to capture the dynamic, subtle, and personalized nature of real-world harassment without the support of dynamic knowledge bases or interactive reasoning capabilities.
	\section{DATASETS AND EVALUATION METRICS} \label{sec:4}
	This section describes several widely used social media datasets, including their sources, formats, composition, collection methods, and sizes. Additionally, we provide the corresponding tasks applicable to each dataset.
	Table \ref{tab:datasets} summarizes the general information for the thirteen datasets.
	We then introduce some standard evaluation metrics.
	
	\begin{table*}[t]
		\centering
		\caption{Summary of Datasets for Mental Disorder Detection.}
		\label{tab:datasets}
		{\renewcommand{\arraystretch}{1.5}
			\begin{tabularx}{\textwidth}{
				>{\raggedright}m{2.5cm}>{\raggedright\arraybackslash}m{1.5cm}>{\raggedright\arraybackslash}m{4cm}>{\raggedright\arraybackslash}m{1.5cm}>{\raggedright\arraybackslash}m{6.7cm}}
				\hline
				\textbf{Dataset} & \textbf{Source} & \textbf{Task} & \textbf{Type} &\textbf{Dataset Size} \\
				\hline
				UMD Suicidality Dataset\cite{chim2024overview} & Reddit & Four-level Depression Detection & Text & 934 Users (Selected from 11,129 users) \\
				\hline
				Dreaddit\cite{turcan2019dreaddit} & Reddit & Binary Stress Prediction & Text & 3553 Posts (52.3\% True, 47.7\% False) \\
				\hline
				DepSeverity\cite{naseem2022early} & Reddit & Four-level Depression Prediction & Text & 3553 Posts (72.9\% Minimum, 8.2\% Mild, \newline 11.3\% Moderate, 7.9\% Severe)\\
				\hline
				STRD\cite{li2022suicide} & Reddit & Four-level Suicide Risk Prediction & Text & Users: 500 (25.8\% Indicator, 40\% Ideation, 26.4\% Behavior, 7.8\% Attempt) \\
				\hline
				SDCNL\cite{haque2021deep} & Reddit & Binary Suicide Risk Prediction & Text & 1895 Posts (48.3\% low, 51.7\% high) \\
				\hline
				CSSRS-Suicide\cite{gaur2019knowledge} & Reddit & Five-level Suicide Risk Prediction & Text & 500 Users (21.6\% Supportive, 19.8\% Indicator, 34.2\% Ideation, 15.4\% Behavior, 9.0\% Attempt) \\
				\hline
				RSDD\cite{yates2017depression} & Reddit  & Binary Depression Prediction & Text & 116,484 Posts (7.9\% True, 92.1\% False) \\
				\hline
				SWMH\cite{ji2021suicidal} & Reddit & Classification of Four Mental Disorders & Text & 54,412 Posts (34.5\% Depression, 18.7\% Suicidal, 17.6\% Anxiety, 14.1\% Bipolar, 15.2\% Offmychest) \\ 
				\hline
				Twt-60Users\cite{jamil2017monitoring} & Twitter (X) & Binary Depression Prediction & Text & 8135 Posts (9.3\% True, 90.7\% False) \\
				\hline
				TMDD\cite{gui2019cooperative} & Twitter (X) & Binary Depression Prediction & Text, Image & Users: 2804 (50\% True, 50\% False) \newline Posts: 1,111,920 (20.9\% True, 79.1\% False) \\
				\hline
				SOS-HL-1K\cite{qi2023supervised} & Weibo & Binary Suicide Risk Prediction & Text & 1249 Posts (51.9\% low, 48.1\% high) \\
				\hline
				SWDD\cite{cai2023depression} & Weibo & Binary Depression Prediction & Text, Image & Users: 23,237 (16\% True, 84\% False) \newline Posts: 4,854,421 (16.2\% True, 83.8\% False) \\
				\hline
				WU3D\cite{wang2020multitask} & Weibo & Binary Depression Prediction & Text, Image & Users: 32,570 (31.7\% True, 68.3\% False) \newline Posts: 2,191,910 (18.7\% True, 81.3\% False)  \\
				\hline
			\end{tabularx}
			}
		\vspace{-5px}
	\end{table*}
	
	\subsection{\textbf{Popular Datasets for Mental Disorders}}
	\textbf{UMD Suicidality Dataset (CLPsych Shared Task)}\cite{chim2024overview}:
	The UMD dataset contains posts and comments from users on Reddit about suicidal intent or behavior.
	The dataset collected 1,556,194 posts from 11,129 users, and after filtering out users with fewer than 10 posts, 934 users were selected for annotation through random sampling.
	
	\textbf{Dreaddit}\cite{turcan2019dreaddit}:
	The dataset is a collection of posts from ten subreddits on Reddit between January 1, 2017 and November 19, 2018 across five domains: social, anxiety, abuse, PTSD, and financial.
	A team of experts independently assessed snippets of posts and subsequently integrated their respective scores to generate final binary labels.
	The dataset is suitable for binary stress detection.
	
	\textbf{DepSeverity}\cite{naseem2022early}:
	The dataset utilized the same posts as Dreaddit, but with a shift in focus to depression content.
	Two experts categorized each post into four depression severity levels based on DSM-5 criteria.
	The dataset was applied to the four levels of depression detection.

	\textbf{STRD}\cite{li2022suicide}:
	This dataset was constructed by collecting posts from subreddits on Reddit between January 2020 and December 2021.
	The final dataset contains 3,998 relevant posts from 1,791 users, manually annotated by 500 random users.
	The annotations include two types of labels: suicide risk level and 17 categories of suicide triggers.
	This dataset is intended for applications such as suicide risk level prediction and suicide trigger detection.
	
	\textbf{SDCNL}\cite{haque2021deep}:
	The dataset collects posts from communities such as r/SuicideWatch and r/Depression via Python Reddit API, covering 1,723 users. Each post was manually reviewed by experts to flag the presence of suicidal ideation.
	The dataset is suitable for binary suicide risk detection.
	
	\textbf{CSSRS-Suicide}\cite{gaur2019knowledge}:
	The CSSRS-Suicide dataset contains posts collected from 15 mental health-related subreddits between 2005 and 2016. 
	Four specialized psychiatrists manually assessed 500 users according to the guidelines of the Columbia Suicide Severity Rating Scale, classifying their suicide risk into five levels.
	The dataset was applied to the five levels of suicide risk detection.
	
	\textbf{RSDD}\cite{yates2017depression}:
	This dataset contains posts from more than 9,000 users who have been diagnosed with depression, as well as posts from more than 107,000 undiagnosed users.
	Importantly, any content from diagnosed users that appeared in mental health-focused subreddits or contained explicit depression-related phrases was excluded from the dataset.
	The dataset was applied to binary depression detection.

	\textbf{SWMH}\cite{ji2021suicidal}:
	This dataset is collected from multiple mental health-related subreddits on Reddit, including r/SuicideWatch.
	This collection contains a total of 54,412 posts.
	It contains discussions about suicide attempts and mental disorders such as depression, anxiety, and bipolar disorder.
	This dataset is used for research on suicidal ideation and multiple mental disorder detection.

	\textbf{Twt-60Users}\cite{jamil2017monitoring}:
	The dataset used the Twitter API to collect tweets from 60 users during 2015.
	The tweets were meticulously annotated by two professionals to determine the presence of depressive signals.
	The dataset is suitable for binary depression detection.
	
	\textbf{TMDD}\cite{gui2019cooperative}:
	The dataset was constructed in two steps. 
	Firstly, tweets from users within a certain time frame were obtained based on self-diagnosis, which constitutes the text depression dataset.
	Subsequently, all images were collected using Twitter API based on the tweet IDs.
	A new multimodal dataset was constructed based on these images and tweets.
	
	\textbf{SOS-HL-1K}\cite{qi2023supervised}:
	The data is obtained by crawling user comments on the blog “Zoufan” in the microblogging platform.
	The dataset consists of textual data, mainly user comments.
	The dataset was annotated by a qualified psychologist.
	The dataset is suitable for binary suicide risk detection.
	
	\textbf{SWDD}\cite{cai2023depression}:
	This dataset contains samples of depressed and non-depressed users.
	The data in this dataset consists of three parts: the user's personal information, the history of tweets, and a symptom description table.
	The dataset is suitable for binary depression detection.
	
	\textbf{WU3D}\cite{wang2020multitask}:
	This dataset includes samples of normal and depressed users.
	It contains user profile information and historical tweets.
	All samples identified as depressed were manually labeled by expert data annotators and subsequently validated by psychologists and psychiatrists.
	The dataset is suitable for binary depression detection.

	\subsection{Evaluation Metrics} \label{Metrics}
	The following are the most commonly used evaluation metrics in current research: Accuracy, Precision, Specificity, Sensitivity/Recall, ROC, PRC, AUC, Kappa statistics, Balanced Accuracy, F1-score, and Weighted F1-Score, with each providing a distinct perspective on model performance.

\begin{itemize}[leftmargin=*]
	\item \textbf{Accuracy (Acc.)}:
	This metric measures the proportion of correct predictions (True Positives + True Negatives) out of all total predictions, providing the most intuitive, general overview of the model's overall correctness.

	\item \textbf{Precision (Prec.)}:
	This metric evaluates the exactness of the model, measuring the proportion of predicted positive instances that were actually positive.

	\item \textbf{Recall (Rec.)}:
	Also known as Sensitivity, this metric assesses the model’s completeness by measuring the proportion of actual positive instances that are correctly identified. 

	\item \textbf{F1-score}:
	This metric calculates the harmonic mean of Precision and Recall, providing a single score that balances the trade-off between minimizing false positives (Precision) and minimizing false negatives (Recall).

	\item \textbf{Balanced Accuracy (B-Acc)}:
	This metric assigns equal weight to both minority and majority classes by calculating the arithmetic mean of the recall for each class, making it particularly useful for imbalanced datasets.

	\item \textbf{Weighted F1-Score (W-F1)}:
	It calculates the F1-score for each class and then takes a weighted average across classes, making it particularly useful for imbalanced datasets.
	Similarly, there are other variations of the F1-score, such as Macro and Micro F1-Score.
\end{itemize}

	While these metrics are often used, they may not always reflect true performance, especially in the high-stakes and sensitive domain of mental health.
	We primarily recommend the use of Balanced Accuracy and Weighted F1-Score as core evaluation metrics due to their robustness to class imbalance compared to standard accuracy or F1-Score.
	As observed in the experimental datasets, the proportion of samples indicating an illness is often significantly smaller than that of the healthy population. 
	In such scenarios, a naive model that predicts every sample as ``healthy" could achieve a very high standard accuracy score, yet it would offer no clinical value.
	In Section~\ref{sec:5}, we will also use these two evaluation metrics.

\begin{figure*}[t]
	\centering
	\includegraphics[width=\textwidth]{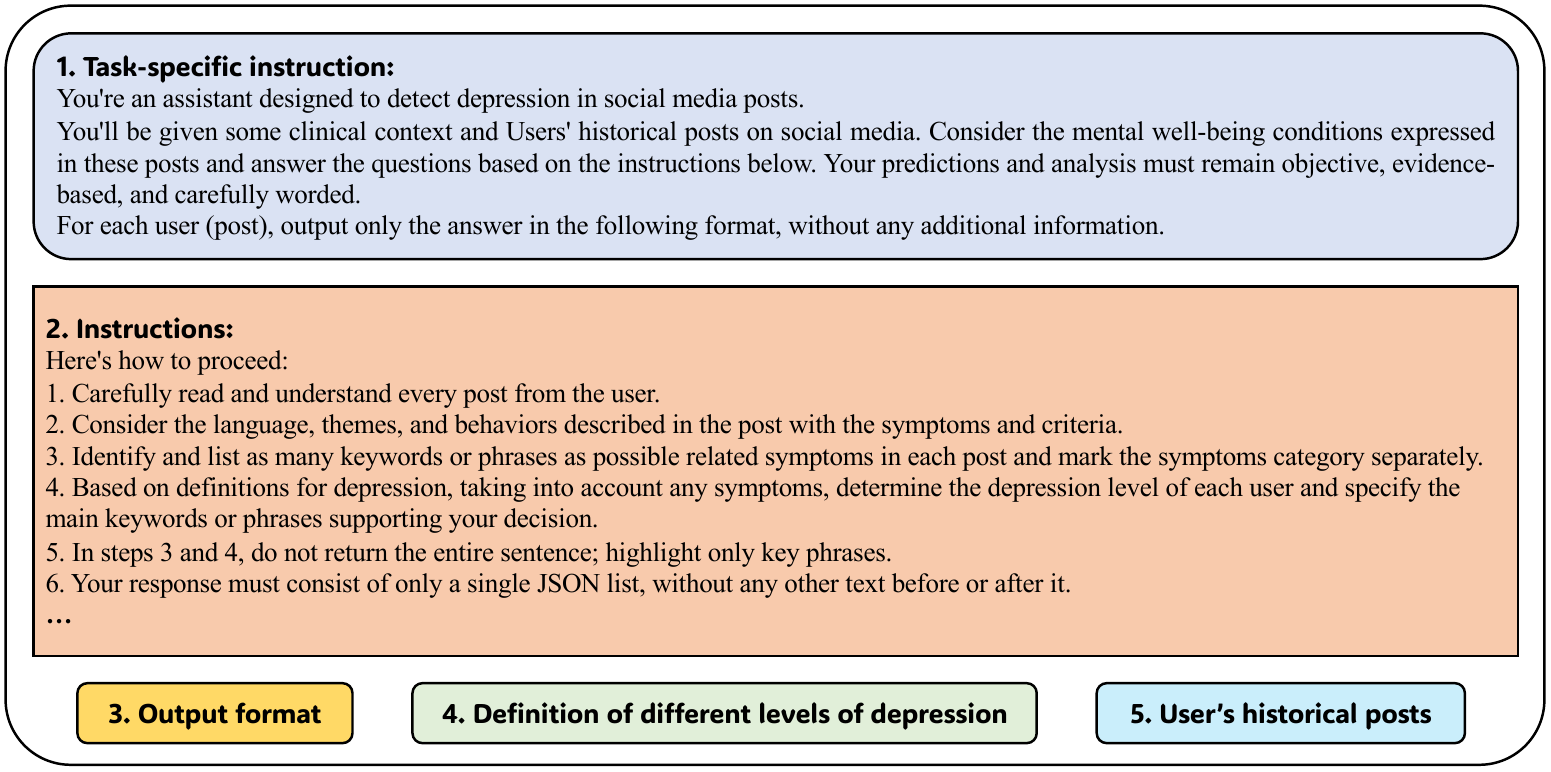}
	\caption{A Task-specific Prompt Example for Identifying Depression Risk in User Posts. The prompt is structured into five distinct modules that define the assistant's role, outline step-by-step reasoning instructions, and specify strict output constraints. It incorporates clinical definitions of depression levels and user historical posts to guide the model in generating objective, evidence-based diagnoses in a structured JSON format.}
	\label{fig:fig9}
	\vspace{-10px}
\end{figure*}
	\section{Empirical Studies} \label{sec:5}
In this section, we conduct empirical studies to evaluate the performance and explainable capability of various LLMs in detecting mental disorders on social media. 
First, we focus on three main approaches: Zero-shot Prompting~\ref{sec:zero_shot}, Few-shot Prompting~\ref{sec:few_shot}, and RAG methods~\ref{subsec:experiments}. 
Then, a case study on explanatory and reasoning abilities is presented~\ref{sec:case}.

\subsection{Experimental Setup}
\subsubsection{\textbf{Model Selection}}
To comprehensively evaluate the capabilities of LLMs in detecting mental disorders on social media, we selected seven models ranging in size and architecture: GPT-4o~\cite{hurst2024gpt}, DeepSeek-R1~\cite{guo2025deepseek}, DeepSeek-V3.1~\cite{liu2024deepseekv3}, Qwen3-32B~\cite{yang2025qwen3}, Llama-3.3-70B, Llama-3.1-70B, and Llama-3.1-405B~\cite{dubey2024llama}.

\subsubsection{\textbf{Datasets and Tasks}}
We utilized six datasets, namely \textit{CSSRS-S}, \textit{STRD}, \textit{SDCNL}, \textit{DepSeverity}, \textit{Dreaddit}, and \textit{SWMH}, encompassing both binary and multi-class classification tasks to test the models' granularity in symptom detection. Details are shown in Table~\ref{tab:datasets2}.

\begin{table}[t]
    \centering
    \caption{Overview of Datasets, Target Conditions, and Classification Granularity}
    \label{tab:datasets2}
    \renewcommand{\arraystretch}{1.2} 
    \begin{tabular}{lllc}
        \toprule
        \textbf{Category} & \textbf{Dataset} & \textbf{Task Granularity} & \textbf{Avg.T.} \\
        \midrule
        \multirow{3}{*}{\textbf{Suicide Risk}} 
            & CSSRS-S\cite{gaur2019knowledge} 		& 5-level Classification & 16 \\
            & STRD\cite{li2022suicide}          & 4-level Classification & 2 \\
            & SDCNL\cite{haque2021deep}         & Binary Classification  & 2 \\
        \midrule
        \textbf{Depression} & DepSeverity\cite{naseem2022early} & 4-level Classification & 1 \\
        \midrule
        \textbf{Stress} & Dreaddit\cite{turcan2019dreaddit} & Binary Classification & 1 \\
        \midrule
        \textbf{Anxiety} & SWMH\cite{ji2021suicidal} & Binary Classification & 2 \\
		\midrule
        \textbf{Bipolar} & SWMH & Binary Classification & 2 \\
        \bottomrule
		\multicolumn{4}{p{0.96\linewidth}}{\footnotesize
        \textbf{Note.} Avg.T. denotes the \emph{average token} length of a raw example in each dataset, computed by directly tokenizing each entry with the backbone LLM tokenizer. It excludes tokens from prompt templates or few-shot demonstrations and is reported in \emph{units of 100 tokens.}}
    \end{tabular}
\vspace{-10px}
\end{table}

\subsubsection{\textbf{Evaluation Metrics}}
Following the analysis in Section~\ref{Metrics}, we adopt Balanced Accuracy (B-Acc) and Weighted F1-Score (W-F1) as metrics to evaluate model performance.

\subsection{Zero-shot Prompting Experiment Results and Analysis} \label{sec:zero_shot}
\subsubsection{\textbf{Experiment Settings}}
For our zero-shot evaluation, we employed an enhanced prompting strategy rather than a simple query, providing the LLM with comprehensive instructions on how to process the task. 
As illustrated in Figure~\ref{fig:fig9}, this structured prompt consists of five key components: (i) a task-specific instruction defining the model's role and objective, (ii) detailed instructions outlining the step-by-step reasoning, (iii) the required output format, (iv) clinical definitions for the different levels of the [target disorder], and (v) the user's historical posts.

\subsubsection{\textbf{Overall Performance}}
Results are shown in Table~\ref{tab:experiment_results}.

\begin{table*}[htbp]
\centering
\caption{Experiment results under zero-shot and few-shot prompting}
\label{tab:experiment_results}
\begin{adjustbox}{max width=\textwidth}
\begin{threeparttable}
\setlength{\tabcolsep}{6pt}
\begin{tabular}{c|l|cc|cc|cc|cc|cc|cc|cc}
\hline
\multicolumn{2}{c|}{\textbf{Datasets}} &
\multicolumn{2}{c|}{\textbf{CSSRS-Suicide}} &
\multicolumn{2}{c|}{\textbf{SDCNL}} &
\multicolumn{2}{c|}{\textbf{STRD}} &
\multicolumn{2}{c|}{\textbf{DepSeverity}} &
\multicolumn{2}{c|}{\textbf{Dreaddit}} &
\multicolumn{2}{c|}{\textbf{SWMH-Anxiety}} &
\multicolumn{2}{c}{\textbf{SWMH-Bipolar}} \\
\cdashline{1-16}
\textbf{Settings} & \textbf{Models} &
B-Acc & W-F1 & B-Acc & W-F1 & B-Acc & W-F1 & B-Acc & W-F1 & B-Acc & W-F1 & B-Acc & W-F1 & B-Acc & W-F1 \\
\hline
\multirow{7}{*}{\rotatebox{90}{\textbf{Zero-shot Prompt}}}
& GPT-4o & 0.473 & 0.434 & \textbf{0.710} & \textbf{0.706} & 0.676 & 0.631 & 0.396 & 0.610 & 0.733 & 0.731 & 0.611 & 0.786 & 0.753 & 0.644 \\
& Qwen3-32B & 0.511 & \textbf{0.494} & 0.676 & 0.672 & 0.652 & 0.613 & \textbf{0.411} & \textbf{0.662} & \textbf{0.796} & \textbf{0.798} & \textbf{0.628} & 0.777 & 0.729 & 0.623 \\
& DeepSeek-R1 & \textbf{0.549} & 0.485 & 0.695 & 0.694 & \textbf{0.699} & \textbf{0.634} & 0.408 & 0.645 & 0.764 & 0.765 & 0.599 & 0.791 & 0.755 & 0.662 \\
& DeepSeek-V3.1 & 0.434 & 0.367 & 0.704 & 0.700 & 0.649 & 0.603 & 0.403 & 0.566 & 0.719 & 0.711 & 0.574 & 0.783 & 0.775 & 0.710 \\
& Llama-3.3-70B & 0.438 & 0.465 & 0.616 & 0.582 & 0.610 & 0.523 & 0.374 & 0.417 & 0.674 & 0.653 & 0.569 & 0.784 & 0.793 & 0.731 \\
& Llama-3.1-405B & 0.381 & 0.426 & 0.685 & 0.676 & 0.602 & 0.593 & 0.347 & 0.443 & 0.745 & 0.744 & 0.610 & \textbf{0.796} & \textbf{0.808} & \textbf{0.765} \\
& Llama-3.1-70B & 0.376 & 0.418 & 0.635 & 0.599 & 0.596 & 0.526 & 0.391 & 0.477 & 0.669 & 0.648 & \textbf{0.628} & 0.793 & 0.779 & 0.734 \\
\hline
\multirow{7}{*}{\rotatebox{90}{\textbf{Few-shot Prompt}}}
& GPT-4o & 0.501 & 0.458 & \textbf{0.714} & \textbf{0.711} & 0.648 & 0.604 & 0.381 & 0.542 & \textbf{0.791} & \textbf{0.793} & 0.619 & 0.756 & 0.763 & 0.672 \\
& Qwen3-32B & 0.509 & \textbf{0.462} & 0.697 & 0.696 & 0.649 & 0.587 & \textbf{0.405} & \textbf{0.650} & 0.759 & 0.750 & \textbf{0.684} & 0.767 & 0.736 & 0.629 \\
& DeepSeek-R1 & \textbf{0.534} & 0.438 & 0.704 & 0.704 & \textbf{0.702} & \textbf{0.622} & 0.395 & 0.589 & 0.782 & 0.783 & \textbf{0.611} & \textbf{0.782} & 0.776 & 0.694 \\
& DeepSeek-V3.1 & 0.470 & 0.406 & 0.706 & 0.704 & 0.626 & 0.577 & 0.381 & 0.501 & 0.753 & 0.751 & 0.588 & 0.778 & 0.800 & 0.750 \\
& Llama-3.3-70B & 0.450 & 0.447 & 0.668 & 0.658 & 0.572 & 0.469 & 0.351 & 0.389 & 0.716 & 0.706 & 0.555 & 0.769 & 0.798 & 0.734 \\
& Llama-3.1-405B & 0.395 & 0.451 & 0.688 & 0.654 & \textbf{0.642} & \textbf{0.635} & 0.340 & 0.393 & 0.780 & 0.785 & 0.607 & 0.770 & 0.794 & 0.723 \\
& Llama-3.1-70B & 0.429 & 0.407 & 0.666 & 0.657 & 0.517 & 0.411 & 0.346 & 0.360 & 0.716 & 0.707 & 0.553 & 0.765 & \textbf{0.803} & \textbf{0.755} \\
\hline
\end{tabular}
	\begin{tablenotes}[flushleft]
		\footnotesize
        \item \textbf{Note.} We report balanced accuracy (B-Acc) and weighted F1 (W-F1) for each model and dataset under zero-shot and few-shot prompting settings. 
		Higher values indicate better performance. 
		Bold numbers denote the best performance for each dataset and metric across all models.
    \end{tablenotes}
\end{threeparttable}
\end{adjustbox}
\vspace{-10px}
\end{table*}

\textbf{Disease and Task Perspective.} 
Task complexity acts as the primary determinant of performance, where models excel in binary classifications but falter significantly in fine-grained multi-class scenarios, regardless of the specific pathology.
For instance, on the binary \textit{Dreaddit} and \textit{SDCNL} datasets, top models like Qwen3-32B achieve W-F1 scores as high as 0.798 and 0.672, respectively. In contrast, efficacy drops precipitously when models face multi-class problems requiring fine-grained differentiation. 
On the four-level \textit{STRD} task, the same Qwen3-32B model drops to a W-F1 of 0.613, and on the five-level \textit{CSSRS}, the highest W-F1 across all models is only 0.494 (Qwen3-32B). Additionally, the \textit{SWMH} datasets highlight diagnostic bias: on \textit{SWMH-Anxiety}, Llama-3.1-405B achieves a high W-F1 of 0.796 but a significantly lower B-Acc of 0.610, indicating a tendency to over-predict the majority ``anxious" class due to broad diagnostic criteria.

\textbf{Dataset Characteristics (Length and Labels).} 
Performance is intrinsically linked to data granularity and context availability, with the intersection of sparse context and multi-level labeling presenting the most significant hurdle for zero-shot inference.
The combination of short text and multi-level labels proves most challenging. The \textit{DepSeverity} dataset yields poor results, with Llama-3.1-405B achieving a W-F1 of only 0.443, far below its binary classification scores. 
This confirms that without sufficient context, LLMs struggle to resolve subtle severity boundaries. Conversely, extreme length introduces its own challenges but allows reasoning-heavy models to shine. 
On \textit{CSSRS}, while overall W-F1 is low, the reasoning-oriented DeepSeek-R1 achieves the highest B-Acc of 0.549, suggesting that longer contexts allow specialized models to better identify minority classes despite the difficulty of the 5-level schema.

\textbf{Model Size and Architecture.} 
Architectural specialization and parameter scaling are decisive factors, as reasoning-enhanced models dominate complex tasks while larger parameter counts drive superior generalization across standard architectures.
First, reasoning-optimized models outperform general-purpose ones on complex tasks. DeepSeek-R1 dominates the hardest tasks, achieving the top B-Acc on both \textit{CSSRS} (0.549) and \textit{STRD} (0.699), surpassing GPT-4o (0.473 and 0.676, respectively). 
Second, within the same family, scaling parameters enhance generalization. 
The Llama-3.1-405B significantly outperforms its 70B counterpart; for example, on \textit{SWMH-Bipolar}, the 405B model achieves a B-Acc of 0.808 compared to the 70B's 0.779.
This validates that for general architectures, increasing parameter count directly translates to better distinctiveness in zero-shot mental health analysis.

\subsection{Few-shot Prompting Experiment Results and Analysis} \label{sec:few_shot}
\definecolor{mygreen}{RGB}{20, 100, 20}
\definecolor{myred}{RGB}{160, 10, 10}
\newcommand{\pmnum}[1]{\IfBeginWith{#1}{-}{\textcolor{myred}{$\downarrow$ #1}}{\textcolor{mygreen}{$\uparrow$ #1}}}

\subsubsection{\textbf{Experiment Settings}}
\begin{figure}[t]
	\centering
	\includegraphics[width=0.48\textwidth]{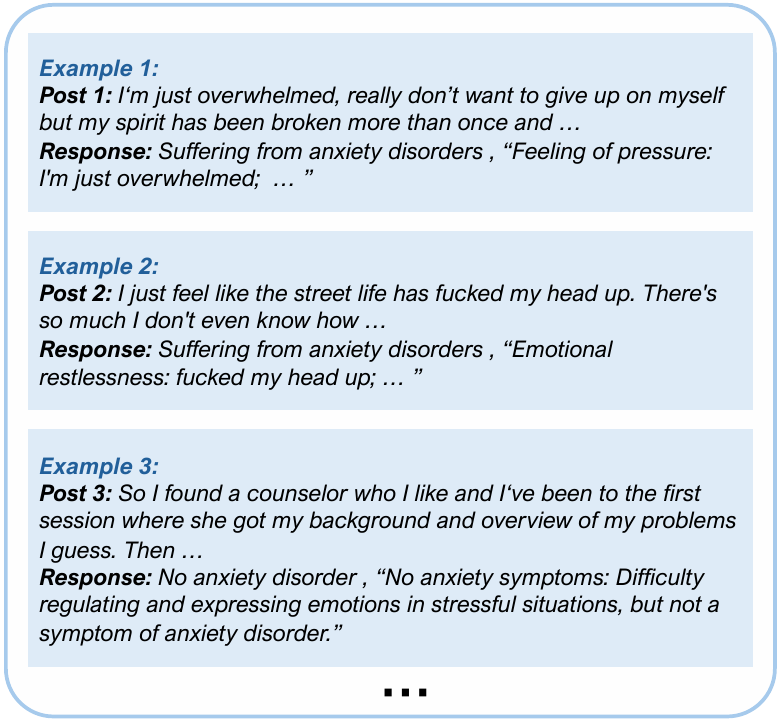}
	\caption{Examples of Post Analysis for Anxiety Disorder Detection with Few-shot Method. All examples presented here are sampled from the Anxiety category within the SWMH dataset.}
	\label{fig:fig10}
	\vspace{-10px}
\end{figure}

For the few-shot experiments, we adapted the zero-shot prompt structure by incorporating in-context examples. 
For each dataset, we estimate the maximum number of few-shot examples that can fit into the context window while reserving appropriate tokens for model outputs.
As depicted in Figure \ref{fig:fig10}, each demonstration is formatted as a (Post, Response) pair, where ‘Post' represents a user's historical posts, and ‘Response' describes the identified symptoms, followed by direct quotations from the ‘Post’ as supporting evidence.
These examples were randomly sampled from our dataset, and their corresponding “Responses” were manually written by three researchers with expertise in mental health. 
The responses were refined through two iterative rounds of review, with each round involving independent assessments by all three researchers to ensure high quality.

\subsubsection{\textbf{Overall Performance}}
The results and changes are shown in Table~\ref{tab:experiment_results} and Table~\ref{tab:performance_delta}, respectively.

\begin{table*}[htbp]
\centering
\caption{Balanced accuracy and weighted F1 change with few-shot prompting}
\label{tab:performance_delta}
\setlength{\tabcolsep}{6.5pt} 
\begin{adjustbox}{max width=\textwidth}
\begin{threeparttable}
\begin{tabular}{ll|ccccccc}
\hline
\textbf{Model} & \textbf{Metric} & \textbf{CSSRS-Suicide} & \textbf{SDCNL} & \textbf{STRD} & \textbf{DepSeverity} & \textbf{Dreaddit} & \textbf{SWMH-Anxiety} & \textbf{SWMH-Bipolar} \\
\hline
\multirow{2}{*}{GPT-4o}
& $\Delta$ W-F1 & \textbf{\pmnum{+0.024}} & \textbf{\pmnum{+0.005}} & \textbf{\pmnum{-0.027}} & \textbf{\pmnum{-0.068}} & \textbf{\pmnum{+0.062}} & \textbf{\pmnum{-0.030}} & \textbf{\pmnum{+0.028}} \\
& $\Delta$ B-Accuracy & \textbf{\pmnum{+0.028}} & \textbf{\pmnum{+0.004}} & \textbf{\pmnum{-0.028}} & \textbf{\pmnum{-0.015}} & \textbf{\pmnum{+0.058}} & \textbf{\pmnum{+0.008}} & \textbf{\pmnum{+0.010}} \\
\cline{1-9}
\multirow{2}{*}{Qwen3-32B}
& $\Delta$ W-F1 & \textbf{\pmnum{-0.032}} & \textbf{\pmnum{+0.024}} & \textbf{\pmnum{-0.026}} & \textbf{\pmnum{-0.012}} & \textbf{\pmnum{-0.048}} & \textbf{\pmnum{-0.010}} & \textbf{\pmnum{+0.006}} \\
& $\Delta$ B-Accuracy & \textbf{\pmnum{-0.002}} & \textbf{\pmnum{+0.021}} & \textbf{\pmnum{-0.003}} & \textbf{\pmnum{-0.006}} & \textbf{\pmnum{-0.037}} & \textbf{\pmnum{+0.056}} & \textbf{\pmnum{+0.007}} \\
\cline{1-9}
\multirow{2}{*}{DeepSeek-R1}
& $\Delta$ W-F1 & \textbf{\pmnum{-0.047}} & \textbf{\pmnum{+0.010}} & \textbf{\pmnum{-0.012}} & \textbf{\pmnum{-0.056}} & \textbf{\pmnum{+0.018}} & \textbf{\pmnum{-0.009}} & \textbf{\pmnum{+0.032}} \\
& $\Delta$ B-Accuracy & \textbf{\pmnum{-0.015}} & \textbf{\pmnum{+0.009}} & \textbf{\pmnum{+0.003}} & \textbf{\pmnum{-0.013}} & \textbf{\pmnum{+0.018}} & \textbf{\pmnum{+0.012}} & \textbf{\pmnum{+0.021}} \\
\cline{1-9}
\multirow{2}{*}{DeepSeek-V3.1}
& $\Delta$ W-F1 & \textbf{\pmnum{+0.039}} & \textbf{\pmnum{+0.004}} & \textbf{\pmnum{-0.026}} & \textbf{\pmnum{-0.065}} & \textbf{\pmnum{+0.040}} & \textbf{\pmnum{-0.005}} & \textbf{\pmnum{+0.040}} \\
& $\Delta$ B-Accuracy & \textbf{\pmnum{+0.036}} & \textbf{\pmnum{+0.002}} & \textbf{\pmnum{-0.023}} & \textbf{\pmnum{-0.022}} & \textbf{\pmnum{+0.034}} & \textbf{\pmnum{+0.014}} & \textbf{\pmnum{+0.025}} \\
\cline{1-9}
\multirow{2}{*}{Llama-3.3-70B}
& $\Delta$ W-F1 & \textbf{\pmnum{-0.018}} & \textbf{\pmnum{+0.076}} & \textbf{\pmnum{-0.054}} & \textbf{\pmnum{-0.028}} & \textbf{\pmnum{+0.053}} & \textbf{\pmnum{-0.015}} & \textbf{\pmnum{+0.003}} \\
& $\Delta$ B-Accuracy & \textbf{\pmnum{+0.012}} & \textbf{\pmnum{+0.052}} & \textbf{\pmnum{-0.038}} & \textbf{\pmnum{-0.023}} & \textbf{\pmnum{+0.042}} & \textbf{\pmnum{-0.014}} & \textbf{\pmnum{+0.005}} \\
\cline{1-9}
\multirow{2}{*}{Llama-3.1-405B}
& $\Delta$ W-F1 & \textbf{\pmnum{+0.025}} & \textbf{\pmnum{-0.022}} & \textbf{\pmnum{+0.042}} & \textbf{\pmnum{-0.050}} & \textbf{\pmnum{+0.041}} & \textbf{\pmnum{-0.026}} & \textbf{\pmnum{-0.042}} \\
& $\Delta$ B-Accuracy & \textbf{\pmnum{+0.014}} & \textbf{\pmnum{+0.003}} & \textbf{\pmnum{+0.040}} & \textbf{\pmnum{-0.007}} & \textbf{\pmnum{+0.035}} & \textbf{\pmnum{-0.003}} & \textbf{\pmnum{-0.014}} \\
\cline{1-9}
\multirow{2}{*}{Llama-3.1-70B}
& $\Delta$ W-F1 & \textbf{\pmnum{-0.011}} & \textbf{\pmnum{+0.058}} & \textbf{\pmnum{-0.115}} & \textbf{\pmnum{-0.117}} & \textbf{\pmnum{+0.059}} & \textbf{\pmnum{-0.028}} & \textbf{\pmnum{+0.021}} \\
& $\Delta$ B-Accuracy & \textbf{\pmnum{+0.053}} & \textbf{\pmnum{+0.031}} & \textbf{\pmnum{-0.079}} & \textbf{\pmnum{-0.045}} & \textbf{\pmnum{+0.047}} & \textbf{\pmnum{-0.075}} & \textbf{\pmnum{+0.024}} \\
\hline
\end{tabular}
\begin{tablenotes}[flushleft]
        \footnotesize
        \item \textbf{Note.}  Each cell reports the change in weighted F1 ($\Delta$ W-F1) or balanced accuracy ($\Delta$ B-Accuracy) when moving from the zero-shot to the few-shot setting (few-shot minus zero-shot).
		Positive values (\pmnum{+}) indicate improvements from few-shot prompting, whereas negative values (\pmnum{-}) indicate performance drops.
		This table is calculated between the zero-shot and the few-shot sections of Table~\ref{tab:experiment_results}.
\end{tablenotes}
\vspace{-5px}
\end{threeparttable}
\end{adjustbox}
\end{table*}

\textbf{Disease and Task Perspective.}
Few-shot prompting introduces a complexity trap: it reliably improves binary detection, but substantially degrades performance on multi-level tasks by injecting bias.
Binary classifications see consistent gains. On \textit{Dreaddit}, GPT-4o achieves a notable improvement ($\Delta$ W-F1 = +0.062), and even smaller models like Llama-3.1-70B improve by +0.059. 
However, for multi-level tasks, few-shot prompting is detrimental. 
The \textit{DepSeverity} task (4-level) sees a universal collapse, with Llama-3.1-70B suffering a massive drop of $\Delta$ W-F1 = -0.117 and GPT-4o dropping by -0.068. 
This indicates that for high-complexity, subjective tasks, few-shot examples limit generalization by introducing bias.

\textbf{Dataset Characteristics (Length and Labels).}
The effectiveness of few-shot examples is constrained by the text-to-label ratio, acting as noise that induces overfitting in short-text scenarios while providing necessary calibration for information-dense, long-text inputs.
For short-text multi-class datasets like \textit{DepSeverity}, the performance degradation is most severe (all models show negative $\Delta$ W-F1), as the limited context causes models to overfit to the specific phrasing of the few-shot examples. 
In contrast, on the long-text \textit{CSSRS} dataset, few-shot prompting helps calibration. 
Despite the task difficulty, GPT-4o improves its B-Acc by +0.028 and W-F1 by +0.024. 
This suggests that when text is sufficiently long, examples help the model navigate the information density, whereas in short texts, they act as noise.

\textbf{Model Size and Architecture.}
Model sensitivity to few-shot prompting varies by type. 
The reasoning model (DeepSeek-R1) reacts negatively to few-shot injection on complex tasks, with its W-F1 dropping by -0.047 on \textit{CSSRS} and -0.012 on \textit{STRD}. 
Conversely, general LLMs show better adaptability. GPT-4o proves the most robust, recording the highest gains on \textit{Dreaddit} and maintaining positive $\Delta$ on most binary tasks. 
Furthermore, the Llama-3.1-405B demonstrates that scale aids stability; on \textit{SDCNL}, it maintains a positive alignment ($\Delta$ B-Acc +0.003), whereas smaller models exhibit higher variance. 
This reinforces the conclusion that larger parameter sizes provide a buffer against the potential noise introduced by few-shot examples.

\subsection{RAG Experiment Results and Analysis} \label{subsec:experiments}
In this subsection, we present a comprehensive analysis of the experimental results from evaluating baseline few-shot prompting and RAG variants on social media datasets for mental disorder detection.

\subsubsection{\textbf{Experiment Settings}}
We implemented different types of RAG systems, specifically a conventional NaiveRAG based on text chunks and advanced GraphRAG based on knowledge graphs. 
For the GraphRAG implementation, we utilized the LightRAG\cite{guo2024lightrag} and HippoRAG\cite{jimenez2024hipporag} frameworks. 

\textbf{Construction and Retrieval.}
The foundation of our system is a highly domain-specific knowledge base, meticulously curated from authoritative sources. 
This corpus is composed of clinical materials, professional medical literature, and recognized diagnostic standards (such as relevant sections from the DSM-5/ICD-11), ensuring that the information retrieved is accurate, professional, and current, and thus provides a reliable factual grounding for the model's reasoning. 
To facilitate efficient and structured knowledge retrieval, we constructed a knowledge graph over this corpus. 
We leveraged the Qwen3-32B model to perform key information extraction and relation extraction, and then employed the state-of-the-art \texttt{bge-m3}\cite{chen2024bge} model to encode entities and relations into high-dimensional vector embeddings, enabling scalable similarity search over mental health concepts, symptoms, and diagnostic criteria. 
The input text for classification is used as a query to retrieve top-$k$ relevant nodes/snippets from the knowledge graph via vector search, followed by re-ranking with \texttt{bge-reranker-v2-m3} to refine relevance. 
The highest-scoring knowledge snippets are then concatenated with the original task instructions, forming an augmented prompt that is finally fed into the LLM for prediction.

\subsubsection{\textbf{Overall Performance Comparison Across Methods}}
The results are summarized in Table~\ref{tab:comprehensive_results}, evaluating Qwen3-32B (few-shot baseline), NaiveRAG, LightRAG, and HippoRAG.

\begin{table*}[htbp]
\centering
\caption{Experiment results comparison: Baseline vs. Chunk-based and Graph-based RAG approaches}
\label{tab:comprehensive_results}
\setlength{\tabcolsep}{5pt}
\begin{adjustbox}{max width=\textwidth}
\begin{threeparttable}
\begin{tabular}{l|cc|cc|cc|cc|cc|cc|cc}
\hline
\textbf{Datasets} &
\multicolumn{2}{c|}{\textbf{CSSRS-Suicide}} &
\multicolumn{2}{c|}{\textbf{SDCNL}} &
\multicolumn{2}{c|}{\textbf{STRD}} &
\multicolumn{2}{c|}{\textbf{DepSeverity}} &
\multicolumn{2}{c|}{\textbf{Dreaddit}} &
\multicolumn{2}{c|}{\textbf{SWMH-Anxiety}} &
\multicolumn{2}{c}{\textbf{SWMH-Bipolar}} \\
\cdashline{1-15}
\textbf{Methods} &
B-Acc & W-F1 & B-Acc & W-F1 & B-Acc & W-F1 & B-Acc & W-F1 & B-Acc & W-F1 & B-Acc & W-F1 & B-Acc & W-F1 \\
\hline
Qwen3-32B (base) & 0.509 & 0.462 & \textbf{0.697} & \textbf{0.696} & \textbf{0.649} & 0.587 & \textbf{0.405} & \textbf{0.650} & \textbf{0.759} & \textbf{0.750} & \textbf{0.684} & 0.767 & \textbf{0.736} & \textbf{0.629} \\
NaiveRAG (Chunk) & 0.517 & 0.497 & 0.644 & 0.626 & 0.640 & 0.584 & 0.393 & 0.636 & 0.725 & 0.730 & 0.665 & 0.763 & 0.722 & 0.627 \\
LightRAG (Graph) & 0.527 & 0.517 & 0.661 & 0.655 & 0.642 & \textbf{0.592} & 0.396 & 0.639 & 0.737 & 0.734 & 0.657 & 0.765 & 0.727 & 0.616 \\
HippoRAG (Graph) & \textbf{0.529} & \textbf{0.522} & 0.656 & 0.664 & 0.644 & 0.589 & 0.401 & 0.647 & 0.733 & 0.736 & 0.662 & \textbf{0.771} & 0.729 & 0.621 \\
\hline
\end{tabular}
\begin{tablenotes}[flushleft]
        \footnotesize
        \item \textbf{Note.} Comprehensive performance comparison between the baseline model (Qwen3-32B) and three retrieval-augmented generation strategies (NaiveRAG, LightRAG, and HippoRAG) across seven tasks.
        Bold values indicate the best results.
\end{tablenotes}
\vspace{-5px}
\end{threeparttable}
\end{adjustbox}
\end{table*}

In our experiments, the impact of RAG on model performance demonstrated a high degree of task dependency, with its efficacy being closely correlated with the contextual richness of the input text. 
Across the seven evaluated tasks, RAG exhibited definitive performance gains exclusively on the \textit{CSSRS-Suicide} dataset. 
Specifically, NaiveRAG yielded improvements of 0.008 (1.6\%) and 0.035 (7.6\%) in balanced accuracy and W-F1, respectively. 
Knowledge graph-based approaches, or GraphRAG, showed a more pronounced advantage: LightRAG enhanced these metrics by 0.018 (3.5\%) and 0.055 (11.9\%), while HippoRAG achieved a significant performance leap of 0.02 (3.9\%) and 0.06 (13.0\%).

\textbf{Long text benefits, while short text carries the risk of ``noise”.}
On the other six datasets, all tested RAG variants led to a discernible degradation in performance. 
The ability of the LLM to leverage this retrieved knowledge is contingent upon the ``signal density" of the source text. 
The records in the \textit{CSSRS-Suicide} dataset average over 1,600 tokens, significantly longer than the sub-300-token examples in the other datasets, which provide richer narrative detail and clearer symptomatic signals for alignment with retrieved clinical knowledge. In contrast, the brevity of the other datasets leads to sparse signals that hinder symptom identification.
Consequently, the retrieved clinical information often acts as noise rather than a useful signal, thereby impairing the model's judgment and causing performance to decline.

\textbf{Graph-based RAG methods outperform traditional chunk-based RAG.}
A conventional ChunkRAG approach might retrieve isolated text fragments about a single symptom, lacking the context of its complex interrelations. 
In contrast, a knowledge graph-based RAG can traverse nodes and relations to retrieve not only symptom definitions but also their links to diagnostic criteria and shared roles across conditions, providing a more structured, holistic context.
Identifying these shared symptoms is critical for both diagnostic accuracy and explainability. 
By recognizing that non-specific symptoms like fatigue or sadness overlap across disorders (e.g., depression and anxiety), an advanced model like GraphRAG is compelled to seek out the subtle, differentiating features crucial for an accurate differential diagnosis, rather than relying on ambiguous, common indicators. 
This capability directly enhances explainability: the model can justify its conclusion by reasoning that despite the presence of common overlapping symptoms, its diagnosis is based on identifying specific, unique indicators (or the absence of others), thereby mirroring clinical logic and increasing the trustworthiness of its diagnostic reasoning.

\subsubsection{\textbf{Applicability Conditions}}
\label{subsubsec:rag-benefits}
RAG is most suitable under specific conditions, particularly when data is long or structured, such as CSSRS-like narratives exceeding short tweets, when tasks require domain expertise like symptom diagnosis needing DSM references, or when datasets are noisy and diverse, leveraging social media's interactivity and dynamics; it is ideal for rare disorders where baseline LLMs lack sufficient training data. 
Conversely, it is less suitable for very short or sparse posts like those in Dreaddit, where retrieval risks adding noise, and for purely sentiment-based tasks with no need for external facts.
High-quality knowledge bases are crucial for preventing bias amplification in sensitive mental health contexts. Future work could explore hybrid approaches, such as an agentic system, to broaden applicability and further improve performance in challenging social media settings.
\vspace{-3px}

\begin{figure*}[t]
	\centering
	\includegraphics[width=0.9\textwidth]{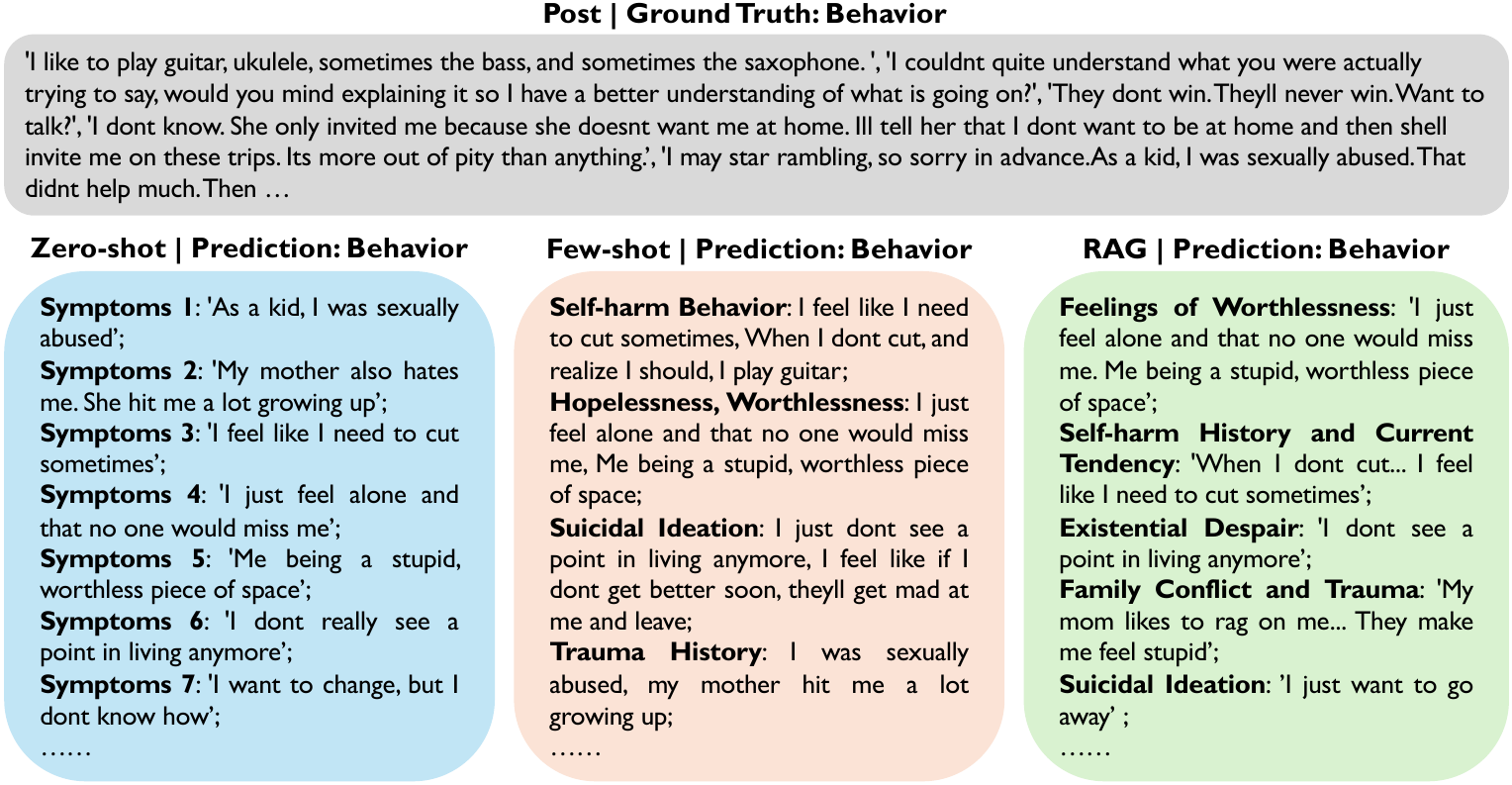}
	\caption{A Case Study of Reasoning Examples on Suicide Detection. Selected from the CSSRS-S dataset, this example illustrates how different methods acquire evidence. The retrieved knowledge enables the model to contextualize the user's distress, resulting in a more accurate risk assessment.}
	\label{fig:fig11}
	\vspace{-10px}
\end{figure*}

\subsection{Case Study of Explainable LLM} \label{sec:case}
Beyond evaluating LLM performance on classification tasks, we conduct a preliminary exploration of their capabilities in mental health explanation.
This capability represents a significant advantage of LLMs, given their ability to generate human-like, natural language responses based on their embedded knowledge.
Given the high cost associated with systematically evaluating explanation outputs, we present several examples as a qualitative case study of different LLMs.
It is important to note that our intention is not to claim that certain LLMs possess superior or inferior explanation abilities. Instead, this section aims to offer a general understanding of how LLMs perform on mental health reasoning tasks.
Specifically, we first conducted a comparative analysis of the Qwen3-32B model's outputs under the Zero-shot, Few-shot, and RAG methods on the CSSRS (suicide risk) dataset, with a primary focus on the quality of the diagnostic symptoms they identified. 
Subsequently, we compared the outputs of GPT-4o, Qwen3-32B, DeepSeek-R1, and Llama-3.3-70B using the RAG method on the DepSeverity (depression) dataset. 

\subsubsection{\textbf{Differences in reasoning ability across methods}}
Figure \ref{fig:fig11} presents a qualitative comparison of the reasoning abilities among the Zero-shot, Few-shot, and RAG methods, based on outputs generated by the Qwen3-32B model for a high-risk post (Ground Truth: Behavior). 
The Zero-shot method correctly identifies the risk but provides only a rudimentary explanation, enumerating extracted phrases (e.g., ``I feel like I need to cut sometimes") as a simple list of symptoms without clinical categorization. 
The Few-shot method demonstrates a more advanced capability by grouping these symptoms into clinically relevant categories (e.g., ``Self-harm behavior"), guided by the in-context examples. 
However, the RAG method provides the most sophisticated and clinically aligned reasoning. 
It not only categorizes symptoms but also introduces more nuanced concepts, such as identifying ``Existential despair" and making a deeper inference of ``Suicidal ideation" from the user's description of thoughts (``I just want to go away"). 
This suggests that by integrating external knowledge, LLMs can move beyond simple symptom matching to a more comprehensive synthesis of the user's psychological state.

\begin{figure}[t]
	\centering
	\includegraphics[width=0.48\textwidth]{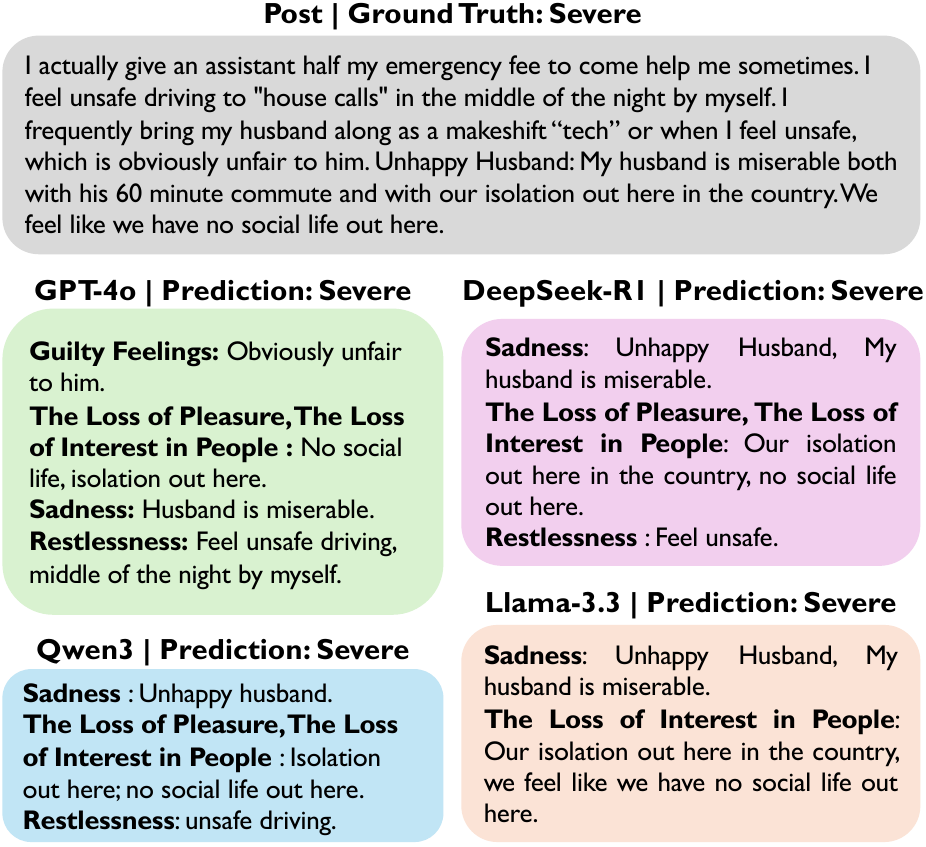}
	\caption{A Case Study of Reasoning Examples on Depression Detection. Derived from the DepSeverity dataset, this case demonstrates the model's ability to identify specific depressive symptoms.}
	\label{fig:fig12}
	\vspace{-10px}
\end{figure}

\subsubsection{\textbf{Diverse Reasoning Capabilities across LLMs}}
To investigate the diverse reasoning capabilities of LLMs when augmented with RAG, we conducted a qualitative analysis on a ``Severe" depression case, focusing on the models' ability to map user-generated text to specific BDI-aligned symptoms, as illustrated in Figure~\ref{fig:fig12}. 
All evaluated models successfully predicted the ``Severe" classification. 
However, significant variations emerged in the depth and comprehensiveness of their explanatory reasoning. 
GPT-4o provided the most thorough analysis, accurately identifying four distinct BDI categories\cite{steer1999common}. 
In contrast, while DeepSeek-R1 and Qwen3 also identified multiple valid symptoms like ``Sadness and Restlessness", they failed to capture the more nuanced symptom of ``Guilty Feelings". 
Llama-3.3 exhibited the most limited reasoning, identifying only ``Sadness" and ``The Loss of Interest in People". 
This case study demonstrates that despite an identical retrieved context, a model's sensitivity to subtle emotional cues is decisive for generating clinically aligned assessments.

	\section{Limitations and Future Directions} \label{sec:6}

\subsection{Multimodal Scarcity and Synthetic Data}
While social media offers rich multimodal cues, existing research predominantly relies on text-based signals. 
The development of large-scale, multimodal benchmarks is severely hindered by strict privacy regulations (e.g., GDPR) and ethical concerns regarding personally identifiable information (face, voiceprint) contained in audio-visual data\cite{hu2021privacy}.

To bypass these privacy constraints, future research should create and collect high-fidelity, anonymized synthetic multimodal datasets. 
This shift enables mental health phenotypic analysis, moving beyond simple labels to quantify nuanced nonverbal markers, such as vocal prosody for psychomotor retardation or facial affect for anhedonia, thereby facilitating precise mental health care without compromising user privacy.
\vspace{-10px}

\subsection{RAG-Driven Clinical Interpretability}
DLMs and PLMs often operate as ``black boxes", lacking the interpretability required for clinical settings~\cite{zhu2025leveraging,garcia2024promoting}. 
Current models frequently fail to align with established diagnostic criteria, leading to high risks of false positives or false negatives without providing verifiable reasoning.

To enhance trustworthiness and clinical alignment, future studies must integrate RAG to ground LLM outputs in external authoritative knowledge~\cite{wang2021context}. 
By dynamically retrieving information from clinical documents, case studies, and knowledge graphs~\cite{jia2023air,wang2024kglink}, LLMs can ensure that diagnostic predictions are supported by cited clinical precedents, effectively mitigating hallucinations and enabling personalized interventions based on patient history.

\subsection{Agentic Reasoning and Forecasting}
Static classification models struggle with implicit forms of abuse and distress, such as sarcasm, memes, or culturally specific expressions. 
Furthermore, current systems are largely confined to simple diagnostic tasks, failing to correlate linguistic signals with high-stakes real-world outcomes, such as acute self-harm risks or public safety threats.

Moving beyond static classification, future research should pivot towards multimodal-agentic systems capable of autonomous planning. 
By dynamically selecting and orchestrating appropriate tools, these agents can effectively address complex scenarios, bridging the gap between precise prediction and proactive intervention.

		\section{CONCLUSION} \label{sec:7}
	In this survey, we present a comprehensive review of LLM-based methods for detecting mental disorders on social media. 
	We summarize recent advances in mental disorder detection, covering various LLMs, RAG techniques, agentic systems, and analyze their strengths and limitations. 
	Our experiments show that, although LLMs excel at understanding complex, unstructured text, their performance remains suboptimal when directly applied to challenging classification and prediction tasks. 
	We outline key directions for future research: integrating multimodal signals for holistic profiling, leveraging RAG to ground trustworthy and faithful explanations, and advancing towards agentic systems capable of autonomous planning.

	\bibliography{references}

@inproceedings{alhamed2024classifying,
  title     = {Classifying Social Media Users before and after Depression Diagnosis via Their Language Usage: A Dataset and Study},
  author    = {Alhamed, Falwah and others},
  booktitle = {Proc. LREC-COLING},
  pages     = {3250--3260},
  year      = {2024}
}

@article{bae2021schizophrenia,
  title     = {Schizophrenia detection using machine learning approach from social media content},
  author    = {Bae, Yi Ji and others},
  journal   = {Sensors},
  volume    = {21},
  number    = {17},
  pages     = {5924},
  year      = {2021}
}

@article{bauer2024using,
  title     = {Using Large Language Models to Understand Suicidality in a Social Media--Based Taxonomy of Mental Health Disorders: Linguistic Analysis of Reddit Posts},
  author    = {Bauer, Brian and others},
  journal   = {JMIR Mental Health},
  volume    = {11},
  pages     = {e57234},
  year      = {2024}
}

@article{bhaumik2023mindwatch,
  title     = {Mindwatch: A smart cloud-based ai solution for suicide ideation detection leveraging large language models},
  author    = {Bhaumik, Runa and others},
  journal   = {MedRxiv},
  year      = {2023}
}

@article{biedermann2016psychotic,
  title     = {Psychotic disorders in DSM-5 and ICD-11},
  author    = {Biedermann, Falko and others},
  journal   = {CNS Spectrums},
  volume    = {21},
  number    = {4},
  pages     = {349--354},
  year      = {2016}
}

@article{birnbaum2017collaborative,
  title     = {A collaborative approach to identifying social media markers of schizophrenia by employing machine learning and clinical appraisals},
  author    = {Birnbaum, Michael L and others},
  journal   = {JMIR},
  volume    = {19},
  number    = {8},
  pages     = {e7956},
  year      = {2017}
}

@article{brown2020language,
  title     = {Language models are few-shot learners},
  author    = {Brown, Tom and Mann, Benjamin and others},
  journal   = {Proc. NeurIPS},
  volume    = {33},
  pages     = {1877--1901},
  year      = {2020}
}

@article{cai2023depression,
  title     = {Depression detection on online social network with multivariate time series feature of user depressive symptoms},
  author    = {Cai, Yicheng and others},
  journal   = {Expert Syst. Appl.},
  volume    = {217},
  pages     = {119538},
  year      = {2023}
}

@article{caspi2020longitudinal,
  title     = {Longitudinal assessment of mental health disorders and comorbidities across 4 decades among participants in the Dunedin birth cohort study},
  author    = {Caspi, Avshalom and others},
  journal   = {JAMA Netw. Open},
  volume    = {3},
  number    = {4},
  pages     = {e203221--e203221},
  year      = {2020}
}

@inproceedings{chim2024overview,
  title     = {Overview of the clpsych 2024 shared task: Leveraging large language models to identify evidence of suicidality risk in online posts},
  author    = {Chim, Jenny and others},
  booktitle = {Proc. CLPsych},
  pages     = {177--190},
  year      = {2024}
}

@article{costello2016early,
  title     = {Early detection and prevention of mental health problems: developmental epidemiology and systems of support},
  author    = {Costello, E Jane},
  journal   = {J. Clin. Child Adolesc. Psychol.},
  volume    = {45},
  number    = {6},
  pages     = {710--717},
  year      = {2016}
}

@article{di2023methodologies,
  title     = {Methodologies for monitoring mental health on Twitter: Systematic review},
  author    = {Di Cara, Nina H and others},
  journal   = {JMIR},
  volume    = {25},
  pages     = {e42734},
  year      = {2023}
}

@article{dubey2024llama,
  title     = {The llama 3 herd of models},
  author    = {Dubey, Abhimanyu and others},
  journal   = {arXiv preprint arXiv:2407.21783},
  year      = {2024}
}

@article{garabiles2019exploring,
  title     = {Exploring comorbidity between anxiety and depression among migrant Filipino domestic workers: a network approach},
  author    = {Garabiles, Melissa R and others},
  journal   = {J. Affect. Disord.},
  volume    = {250},
  pages     = {85--93},
  year      = {2019}
}

@article{garg2023mental,
  title     = {Mental health analysis in social media posts: a survey},
  author    = {Garg, Muskan},
  journal   = {Arch. Comput. Methods Eng.},
  volume    = {30},
  number    = {3},
  pages     = {1819--1842},
  year      = {2023}
}

@inproceedings{gaur2019knowledge,
  title     = {Knowledge-aware assessment of severity of suicide risk for early intervention},
  author    = {Gaur, Manas and others},
  booktitle = {Proc. WWW},
  pages     = {514--525},
  year      = {2019}
}

@inproceedings{gui2019cooperative,
  title     = {Cooperative multimodal approach to depression detection in twitter},
  author    = {Gui, Tao and others},
  booktitle = {Proc. AAAI},
  volume    = {33},
  number    = {01},
  pages     = {110--117},
  year      = {2019}
}

@article{guo2024large,
  title     = {Large language models for mental health applications: Systematic review},
  author    = {Guo, Zhijun and others},
  journal   = {JMIR Mental Health},
  volume    = {11},
  number    = {1},
  pages     = {e57400},
  year      = {2024}
}

@inproceedings{gurumoorthy2019efficient,
  title     = {Efficient data representation by selecting prototypes with importance weights},
  author    = {Gurumoorthy, Karthik S and others},
  booktitle = {Proc. ICDM},
  pages     = {260--269},
  year      = {2019}
}

@article{hadi2023survey,
  title     = {A survey on large language models: Applications, challenges, limitations, and practical usage},
  author    = {Hadi, Muhammad Usman and others},
  journal   = {Authorea Preprints},
  year      = {2023}
}

@inproceedings{haque2021deep,
  title     = {Deep learning for suicide and depression identification with unsupervised label correction},
  author    = {Haque, Ayaan and others},
  booktitle = {Proc. ICANN},
  pages     = {436--447},
  year      = {2021}
}

@article{harrigian2020state,
  title     = {On the state of social media data for mental health research},
  author    = {Harrigian, Keith and others},
  journal   = {arXiv preprint arXiv:2011.05233},
  year      = {2020}
}

@article{hua2024applying,
  title     = {Applying and Evaluating Large Language Models in Mental Health Care: A Scoping Review of Human-Assessed Generative Tasks},
  author    = {Hua, Yining and others},
  journal   = {arXiv preprint arXiv:2408.11288},
  year      = {2024}
}

@article{hua2024large,
  title     = {Large language models in mental health care: a scoping review},
  author    = {Hua, Yining and others},
  journal   = {arXiv preprint arXiv:2401.02984},
  year      = {2024}
}

@phdthesis{jamil2017monitoring,
  title     = {Monitoring tweets for depression to detect at-risk users},
  author    = {Jamil, Zunaira},
  year      = {2017},
  school    = {University of Ottawa}
}

@article{ji2021mentalbert,
  title     = {Mentalbert: Publicly available pretrained language models for mental healthcare},
  author    = {Ji, Shaoxiong and others},
  journal   = {arXiv preprint arXiv:2110.15621},
  year      = {2021}
}

@inproceedings{jia2023air,
  title     = {Air: Adaptive incremental embedding updating for dynamic knowledge graphs},
  author    = {Jia, Zhifeng and Li, Haoyang and Chen, Lei},
  booktitle = {Proc. DASFAA},
  pages     = {606--621},
  year      = {2023}
}

@article{ji2021suicidal,
  title     = {Suicidal ideation and mental disorder detection with attentive relation networks},
  author    = {Ji, Shaoxiong and others},
  journal   = {Neural Comput. Appl.},
  year      = {2021}
}

@article{ke2024exploring,
  title     = {Exploring the frontiers of llms in psychological applications: A comprehensive review},
  author    = {Ke, Luoma and others},
  journal   = {arXiv preprint arXiv:2401.01519},
  year      = {2024}
}

@article{kim2020deep,
  title     = {A deep learning model for detecting mental illness from user content on social media},
  author    = {Kim, Jina and others},
  journal   = {Sci. Rep.},
  volume    = {10},
  number    = {1},
  pages     = {11846},
  year      = {2020}
}

@article{lamichhane2023evaluation,
  title     = {Evaluation of chatgpt for nlp-based mental health applications},
  author    = {Lamichhane, Bishal},
  journal   = {arXiv preprint arXiv:2303.15727},
  year      = {2023}
}

@article{lan2024depression,
  title     = {Depression Detection on Social Media with Large Language Models},
  author    = {Lan, Xiaochong and others},
  journal   = {arXiv preprint arXiv:2403.10750},
  year      = {2024}
}

@article{law2008suicide,
  title     = {Suicide in China: unique demographic patterns and relationship to depressive disorder},
  author    = {Law, Samuel and Liu, Pozi},
  journal   = {Curr. Psychiatry Rep.},
  volume    = {10},
  pages     = {80--86},
  year      = {2008}
}

@article{lawrence2024opportunities,
  title     = {The opportunities and risks of large language models in mental health},
  author    = {Lawrence, Hannah R and others},
  journal   = {JMIR Mental Health},
  volume    = {11},
  number    = {1},
  pages     = {e59479},
  year      = {2024}
}

@article{lee2023rlaif,
  title     = {Rlaif: Scaling reinforcement learning from human feedback with ai feedback},
  author    = {Lee, Harrison and others},
  journal   = {arXiv preprint arXiv:2309.00267},
  year      = {2023}
}

@article{lejeune2022use,
  title     = {Use of social media data to diagnose and monitor psychotic disorders: systematic review},
  author    = {Lejeune, Alban and others},
  journal   = {JMIR},
  volume    = {24},
  number    = {9},
  pages     = {e36986},
  year      = {2022}
}

@article{lewis2020retrieval,
  title     = {Retrieval-augmented generation for knowledge-intensive nlp tasks},
  author    = {Lewis, Patrick and others},
  journal   = {Proc. NeurIPS},
  volume    = {33},
  pages     = {9459--9474},
  year      = {2020}
}

@article{li2022suicide,
  title     = {Suicide risk level prediction and suicide trigger detection: A benchmark dataset},
  author    = {Li, Jun and others},
  journal   = {HKIE Trans.},
  volume    = {29},
  number    = {4},
  pages     = {268--282},
  year      = {2022}
}

@article{li2024survey,
  title     = {A survey on large language model acceleration based on kv cache management},
  author    = {Li, Haoyang and others},
  journal   = {arXiv preprint arXiv:2412.19442},
  year      = {2024}
}

@article{liu2024deepseek,
  title     = {Deepseek-v2: A strong, economical, and efficient mixture-of-experts language model},
  author    = {Liu, Aixin and others},
  journal   = {arXiv preprint arXiv:2405.04434},
  year      = {2024}
}

@article{liu2024enhancing,
  title     = {Enhancing Mental Health Condition Detection on Social Media through Multi-Task Learning},
  author    = {Liu, Jiawen and Su, Menglu},
  journal   = {medRxiv},
  year      = {2024}
}

@article{liu2024visual,
  title     = {Visual instruction tuning},
  author    = {Liu, Haotian and others},
  journal   = {Proc. NeurIPS},
  volume    = {36},
  year      = {2024}
}

@article{lu2024deepseek,
  title     = {Deepseek-vl: towards real-world vision-language understanding},
  author    = {Lu, Haoyu and others},
  journal   = {arXiv preprint arXiv:2403.05525},
  year      = {2024}
}

@article{malhotra2022deep,
  title     = {Deep learning techniques for suicide and depression detection from online social media: A scoping review},
  author    = {Malhotra, Anshu and Jindal, Rajni},
  journal   = {Appl. Soft Comput.},
  volume    = {130},
  pages     = {109713},
  year      = {2022}
}

@article{mcconaughy1993comorbidity,
  title     = {Comorbidity of externalizing and internalizing problems},
  author    = {McConaughy, Stephanie H and Skiba, Russell J},
  journal   = {Sch. Psychol. Rev.},
  volume    = {22},
  number    = {3},
  pages     = {421--436},
  year      = {1993}
}

@article{mcgrath2023age,
  title     = {Age of onset and cumulative risk of mental disorders: a cross-national analysis of population surveys from 29 countries},
  author    = {McGrath, John J and others},
  journal   = {Lancet Psychiatry},
  volume    = {10},
  number    = {9},
  pages     = {668--681},
  year      = {2023}
}

@article{mcmanus2015mining,
  title     = {Mining Twitter data to improve detection of schizophrenia},
  author    = {McManus, Kimberly and others},
  journal   = {AMIA Summits Transl. Sci. Proc.},
  volume    = {2015},
  pages     = {122},
  year      = {2015}
}

@article{metzler2022detecting,
  title     = {Detecting potentially harmful and protective suicide-related content on twitter: machine learning approach},
  author    = {Metzler, Hannah and others},
  journal   = {JMIR},
  volume    = {24},
  number    = {8},
  pages     = {e34705},
  year      = {2022}
}

@article{min2023recent,
  title     = {Recent advances in natural language processing via large pre-trained language models: A survey},
  author    = {Min, Bonan and others},
  journal   = {ACM Comput. Surv.},
  volume    = {56},
  number    = {2},
  pages     = {1--40},
  year      = {2023}
}

@inproceedings{mitchell2015quantifying,
  title     = {Quantifying the language of schizophrenia in social media},
  author    = {Mitchell, Margaret and others},
  booktitle = {Proc. CLPsych},
  pages     = {11--20},
  year      = {2015}
}

@article{mudrik2024exploring,
  title     = {Exploring the role of Large Language Models in haematology: A focused review of applications, benefits and limitations},
  author    = {Mudrik, Aya and others},
  journal   = {Br. J. Haematol.},
  volume    = {205},
  number    = {5},
  pages     = {1685--1698},
  year      = {2024}
}

@inproceedings{naseem2022early,
  title     = {Early identification of depression severity levels on reddit using ordinal classification},
  author    = {Naseem, Usman and others},
  booktitle = {Proc. WWW},
  pages     = {2563--2572},
  year      = {2022}
}

@article{naslund2020social,
  title     = {Social media and mental health: benefits, risks, and opportunities for research and practice},
  author    = {Naslund, John A and others},
  journal   = {J. Technol. Behav. Sci.},
  volume    = {5},
  number    = {3},
  pages     = {245--257},
  year      = {2020}
}

@article{omar2024applications,
  title     = {Applications of Large Language Models in Psychiatry: A Systematic Review.},
  author    = {Omar Sr, Mahmud and others},
  journal   = {medRxiv},
  year      = {2024}
}

@article{omar2024exploring,
  title     = {Exploring the efficacy and potential of large language models for depression: A systematic review},
  author    = {Omar, Mahmud and Levkovich, Inbar},
  journal   = {J. Affect. Disord.},
  year      = {2024}
}

@article{owen2023enabling,
  title     = {Enabling early health care intervention by detecting depression in users of web-based forums using language models: Longitudinal analysis and evaluation},
  author    = {Owen, David and others},
  journal   = {JMIR AI},
  volume    = {2},
  number    = {1},
  pages     = {e41205},
  year      = {2023}
}

@article{owen2024ai,
  title     = {AI for Analyzing Mental Health Disorders Among Social Media Users: Quarter-Century Narrative Review of Progress and Challenges},
  author    = {Owen, David and others},
  journal   = {JMIR},
  volume    = {26},
  pages     = {e59225},
  year      = {2024}
}

@article{qi2023supervised,
  title     = {Supervised Learning and Large Language Model Benchmarks on Mental Health Datasets: Cognitive Distortions and Suicidal Risks in Chinese Social Media},
  author    = {Qi, Hongzhi and others},
  journal   = {arXiv preprint arXiv:2310.15875},
  year      = {2023}
}

@article{qin2023read,
  title     = {Read, diagnose and chat: Towards explainable and interactive LLMs-augmented depression detection in social media},
  author    = {Qin, Wei and others},
  journal   = {arXiv preprint arXiv:2305.05138},
  year      = {2023}
}

@article{radwan2024predictive,
  title     = {Predictive analytics in mental health leveraging llm embeddings and machine learning models for social media analysis},
  author    = {Radwan, Ahmad and others},
  journal   = {IJWSR},
  volume    = {21},
  number    = {1},
  pages     = {1--22},
  year      = {2024}
}

@article{rossler2005size,
  title     = {Size of burden of schizophrenia and psychotic disorders},
  author    = {R{\"o}ssler, Wulf and others},
  journal   = {Eur. Neuropsychopharmacol.},
  volume    = {15},
  number    = {4},
  pages     = {399--409},
  year      = {2005}
}

@inproceedings{sabaneh2023early,
  title     = {Early Risk Prediction of Depression Based on Social Media Posts in Arabic},
  author    = {Sabaneh, Kefaya and others},
  booktitle = {Proc. ICTAI},
  pages     = {595--602},
  year      = {2023}
}

@article{shah2024mental,
  title     = {Mental illness detection through harvesting social media: a comprehensive literature review},
  author    = {Shah, Shahid Munir and others},
  journal   = {PeerJ Comput. Sci.},
  volume    = {10},
  pages     = {e2296},
  year      = {2024}
}

@article{shin2024using,
  title     = {Using large language models to detect depression from user-generated diary text data as a novel approach in digital mental health screening: Instrument validation study},
  author    = {Shin, Daun and others},
  journal   = {JMIR},
  volume    = {26},
  pages     = {e54617},
  year      = {2024}
}

@inproceedings{singh2024extraction,
  title     = {Extraction and summarization of suicidal ideation evidence in social media content using large language models},
  author    = {Singh, Loitongbam Gyanendro and others},
  booktitle = {Proc. CLPsych},
  year      = {2024}
}

@article{skaik2020using,
  title     = {Using social media for mental health surveillance: a review},
  author    = {Skaik, Ruba and Inkpen, Diana},
  journal   = {ACM Comput. Surv.},
  volume    = {53},
  number    = {6},
  pages     = {1--31},
  year      = {2020}
}

@online{social-app-report,
  author = {Business of Apps},
  title  = {Social App Report},
  year   = {2025},
  note   = {Accessed: Feb. 3, 2025},
  url    = {https://www.businessofapps.com/data/social-app-report/}
}

@inproceedings{song2024combining,
  title     = {Combining Hierachical VAEs with LLMs for clinically meaningful timeline summarisation in social media},
  author    = {Song, Jiayu and others},
  booktitle = {Findings ACL},
  pages     = {14651--14672},
  year      = {2024}
}

@article{touvron2023llama,
  title     = {Llama: Open and efficient foundation language models},
  author    = {Touvron, Hugo and others},
  journal   = {arXiv preprint arXiv:2302.13971},
  year      = {2023}
}

@article{turcan2019dreaddit,
  title     = {Dreaddit: A reddit dataset for stress analysis in social media},
  author    = {Turcan, Elsbeth and McKeown, Kathleen},
  journal   = {arXiv preprint arXiv:1911.00133},
  year      = {2019}
}

@article{vaswani2017attention,
  title     = {Attention is all you need},
  author    = {Vaswani, A},
  journal   = {Proc. NeurIPS},
  year      = {2017}
}

@inproceedings{verma2023ai,
  title     = {AI-Enhanced Mental Health Diagnosis: Leveraging Transformers for Early Detection of Depression Tendency in Textual Data},
  author    = {Verma, Srishti and others},
  booktitle = {Proc. ICUMT},
  pages     = {56--61},
  year      = {2023}
}

@article{wang2020depression,
  title     = {Depression risk prediction for chinese microblogs via deep-learning methods: Content analysis},
  author    = {Wang, Xiaofeng and others},
  journal   = {JMIR Med. Inform.},
  volume    = {8},
  number    = {7},
  pages     = {e17958},
  year      = {2020}
}

@article{wang2020multitask,
  title     = {A multitask deep learning approach for user depression detection on sina weibo},
  author    = {Wang, Yiding and others},
  journal   = {arXiv preprint arXiv:2008.11708},
  year      = {2020}
}

@inproceedings{wang2024explainable,
  title     = {Explainable depression detection using large language models on social media data},
  author    = {Wang, Yuxi and others},
  booktitle = {Proc. CLPsych},
  pages     = {108--126},
  year      = {2024}
}

@inproceedings{wang2024kglink,
  title     = {KGLink: A column type annotation method that combines knowledge graph and pre-trained language model},
  author    = {Wang, Yubo and Xin, Hao and Chen, Lei},
  booktitle = {Proc. ICDE},
  pages     = {1023--1035},
  year      = {2024}
}

@online{WHO,
  author = {World Health Organization},
  title  = {WHO Website},
  year   = {2025},
  note   = {Accessed: 2025-02-03},
  url    = {https://www.who.int/}
}

@online{who-mental-disorders,
  author = {{World Health Organization}},
  title  = {Mental Disorders Fact Sheet},
  year   = {2022},
  note   = {Accessed: Oct. 1, 2025},
  url    = {https://www.who.int/news-room/fact-sheets/detail/mental-disorders}
}

@article{wongkoblap2017researching,
  title     = {Researching mental health disorders in the era of social media: systematic review},
  author    = {W., Akkapon and others},
  journal   = {JMIR},
  volume    = {19},
  number    = {6},
  pages     = {e228},
  year      = {2017}
}

@inproceedings{wu2023multimodal,
  title     = {Multimodal large language models: A survey},
  author    = {Wu, Jiayang and others},
  booktitle = {Proc. BigData},
  pages     = {2247--2256},
  year      = {2023}
}

@article{xu2024large,
  title     = {Large language models for generative information extraction: A survey},
  author    = {Xu, Derong and others},
  journal   = {Front. Comput. Sci.},
  volume    = {18},
  number    = {6},
  pages     = {186357},
  year      = {2024}
}

@article{xu2024mental,
  title     = {Mental-llm: Leveraging large language models for mental health prediction via online text data},
  author    = {Xu, Xuhai and others},
  journal   = {ACM IMWUT},
  volume    = {8},
  number    = {1},
  pages     = {1--32},
  year      = {2024}
}

@inproceedings{yang2024mentallama,
  title     = {MentaLLaMA: interpretable mental health analysis on social media with large language models},
  author    = {Yang, Kailai and others},
  booktitle = {Proc. WWW},
  pages     = {4489--4500},
  year      = {2024}
}

@article{yates2017depression,
  title     = {Depression and self-harm risk assessment in online forums},
  author    = {Yates, Andrew and others},
  journal   = {arXiv preprint arXiv:1709.01848},
  year      = {2017}
}

@article{zhang2024mm,
  title     = {Mm-llms: Recent advances in multimodal large language models},
  author    = {Zhang, Duzhen and others},
  journal   = {arXiv preprint arXiv:2401.13601},
  year      = {2024}
}

@article{zhang2024vision,
  title     = {Vision-language models for vision tasks: A survey},
  author    = {Zhang, Jingyi and others},
  journal   = {IEEE TPAMI},
  year      = {2024}
}

@article{kermani2025systematic,
  title     = {A Systematic Evaluation of LLM Strategies for Mental Health Text Analysis: Fine-tuning vs. Prompt Engineering vs. RAG},
  author    = {Kermani, Arshia and others},
  journal   = {arXiv preprint arXiv:2503.24307},
  year      = {2025}
}

@inproceedings{nushida2025automated,
  title     = {An Automated Depression Diagnosis System Utilizing a Knowledge Base Created with GPT},
  author    = {Nushida, Tomoki and others},
  booktitle = {Proc. ICCRD},
  pages     = {329--333},
  year      = {2025}
}

@article{ravenda2025llms,
  title     = {Are llms effective psychological assessors? leveraging adaptive rag for interpretable mental health screening through psychometric practice},
  author    = {Ravenda, Federico and others},
  journal   = {arXiv preprint arXiv:2501.00982},
  year      = {2025}
}

@inproceedings{yao2024personalised,
  title     = {Personalised abusive language detection using llms and retrieval-augmented generation},
  author    = {Yao, Tsungcheng and others},
  booktitle = {Proc. ICNLSP},
  year      = {2024}
}

@inproceedings{antony2025retrieval,
  title     = {Retrieval-Enhanced Mental Health Assessment: Capturing Self-State Dynamics from Social Media Using In-Context Learning},
  author    = {Antony, Anson and Schoene, Annika},
  booktitle = {Proc. CLPsych},
  pages     = {268--278},
  year      = {2025}
}

@inproceedings{xu2024utilizing,
  title     = {Utilizing Large Language Models for Psychological Assessment: Enhancing Suicide Risk Detection Through Social Media Analysis},
  author    = {Xu, Zeling and others},
  booktitle = {Proc. ICFTIC},
  pages     = {1418--1421},
  year      = {2024}
}

@article{gao2023retrieval,
  title     = {Retrieval-augmented generation for large language models: A survey},
  author    = {Gao, Yunfan and others},
  journal   = {arXiv preprint arXiv:2312.10997},
  volume    = {2},
  number    = {1},
  year      = {2023}
}

@inproceedings{wu2025multirag,
  title     = {Multirag: a knowledge-guided framework for mitigating hallucination in multi-source retrieval augmented generation},
  author    = {Wu, Wenlong and others},
  booktitle = {Proc. ICDE},
  pages     = {3070--3083},
  year      = {2025}
}

@article{shuster2021retrieval,
  title     = {Retrieval augmentation reduces hallucination in conversation},
  author    = {Shuster, Kurt and others},
  journal   = {arXiv preprint arXiv:2104.07567},
  year      = {2021}
}

@inproceedings{fan2024survey,
  title     = {A survey on rag meeting llms: Towards retrieval-augmented large language models},
  author    = {Fan, Wenqi and others},
  booktitle = {Proc. KDD},
  pages     = {6491--6501},
  year      = {2024}
}

@inproceedings{fan2025towards,
  title     = {Towards Retrieval-Augmented Large Language Models: Data Management and System Design},
  author    = {Fan, Wenqi and others},
  booktitle = {Proc. ICDE},
  pages     = {4509--4512},
  year      = {2025}
}

@article{ni2025towards,
  title     = {Towards trustworthy retrieval augmented generation for large language models: A survey},
  author    = {Ni, Bo and others},
  journal   = {arXiv preprint arXiv:2502.06872},
  year      = {2025}
}

@inproceedings{yang2025cascadercg,
  title     = {CascadeRCG: Retrieval-Augmented Generation for Enhancing Professionalism and Knowledgeability in Online Mental Health Support},
  author    = {Yang, Di and others},
  booktitle = {Proc. WWW Companion},
  pages     = {1465--1469},
  year      = {2025}
}

@article{amugongo2025retrieval,
  title     = {Retrieval augmented generation for large language models in healthcare: A systematic review},
  author    = {Amugongo, Lameck Mbangula and others},
  journal   = {PLOS Digit. Health},
  volume    = {4},
  number    = {6},
  pages     = {e0000877},
  year      = {2025}
}

@article{lyu2025gpt,
  title     = {Gpt-4v (ision) as a social media analysis engine},
  author    = {Lyu, Hanjia and others},
  journal   = {ACM TIST},
  volume    = {16},
  number    = {3},
  pages     = {1--54},
  year      = {2025}
}

@article{liu2023visual,
  title     = {Visual instruction tuning},
  author    = {Liu, Haotian and others},
  journal   = {Proc. NeurIPS},
  volume    = {36},
  pages     = {34892--34916},
  year      = {2023}
}

@article{yang2025qwen3,
  title     = {Qwen3 technical report},
  author    = {Yang, An and others},
  journal   = {arXiv preprint arXiv:2505.09388},
  year      = {2025}
}

@article{guo2025deepseek,
  title     = {Deepseek-r1: Incentivizing reasoning capability in llms via reinforcement learning},
  author    = {Guo, Daya and others},
  journal   = {arXiv preprint arXiv:2501.12948},
  year      = {2025}
}

@article{hurst2024gpt,
  title     = {Gpt-4o system card},
  author    = {Hurst, Aaron and others},
  journal   = {arXiv preprint arXiv:2410.21276},
  year      = {2024}
}

@article{achiam2023gpt,
  title     = {Gpt-4 technical report},
  author    = {Achiam, Josh and others},
  journal   = {arXiv preprint arXiv:2303.08774},
  year      = {2023}
}

@article{guo2024lightrag,
  title     = {Lightrag: Simple and fast retrieval-augmented generation},
  author    = {Guo, Zirui and others},
  journal   = {arXiv preprint arXiv:2410.05779},
  year      = {2024}
}

@article{jimenez2024hipporag,
  title     = {Hipporag: Neurobiologically inspired long-term memory for large language models},
  author    = {Jimenez Gutierrez, Bernal and others},
  journal   = {Proc. NeurIPS},
  volume    = {37},
  pages     = {59532--59569},
  year      = {2024}
}

@article{chen2024bge,
  title     = {Bge m3-embedding: Multi-lingual, multi-functionality, multi-granularity text embeddings through self-knowledge distillation},
  author    = {Chen, Jianlv and others},
  journal   = {arXiv preprint arXiv:2402.03216},
  year      = {2024}
}

@article{steer1999common,
  title     = {Common and specific dimensions of self-reported anxiety and depression: the BDI-II versus the BDI-IA},
  author    = {Steer, Robert A and others},
  journal   = {Behav. Res. Ther.},
  volume    = {37},
  number    = {2},
  pages     = {183--190},
  year      = {1999}
}

@article{liu2024deepseekv3,
  title     = {Deepseek-v3 technical report},
  author    = {Liu, Aixin and others},
  journal   = {arXiv preprint arXiv:2412.19437},
  year      = {2024}
}

@article{chen2021social,
  title     = {Social media use for health purposes: systematic review},
  author    = {Chen, Junhan and others},
  journal   = {JMIR},
  volume    = {23},
  number    = {5},
  pages     = {e17917},
  year      = {2021}
}

@article{abkenar2021big,
  title     = {Big data analytics meets social media: A systematic review of techniques, open issues, and future directions},
  author    = {Abkenar, Sepideh Bazzaz and others},
  journal   = {Telemat. Inform.},
  volume    = {57},
  pages     = {101517},
  year      = {2021}
}

@article{latkin2017relationship,
  title     = {The relationship between social desirability bias and self-reports of health, substance use, and social network factors among urban substance users in Baltimore, Maryland},
  author    = {Latkin, Carl A and others},
  journal   = {Addict. Behav.},
  volume    = {73},
  pages     = {133--136},
  year      = {2017}
}

@article{van2016validation,
  title     = {Validation of the self reporting questionnaire 20-item (SRQ-20) for use in a low-and middle-income country emergency centre setting},
  author    = {van der Westhuizen, Claire and others},
  journal   = {Int. J. Ment. Health Addict.},
  volume    = {14},
  number    = {1},
  pages     = {37--48},
  year      = {2016}
}

@article{alfonsson2014interformat,
  title     = {Interformat reliability of digital psychiatric self-report questionnaires: a systematic review},
  author    = {Alfonsson, Sven and others},
  journal   = {JMIR},
  volume    = {16},
  number    = {12},
  pages     = {e3395},
  year      = {2014}
}

@article{huang2025survey,
  title     = {A survey on hallucination in large language models: Principles, taxonomy, challenges, and open questions},
  author    = {Huang, Lei and others},
  journal   = {ACM TOIS},
  volume    = {43},
  number    = {2},
  pages     = {1--55},
  year      = {2025}
}

@article{zhang2023large,
  title     = {How do large language models capture the ever-changing world knowledge? a review of recent advances},
  author    = {Zhang, Zihan and others},
  journal   = {arXiv preprint arXiv:2310.07343},
  year      = {2023}
}

@article{chun2025gambling,
  title     = {Gambling-specific metacognitions, depression, and responsible gambling in Macao, China},
  author    = {Chun, FENG and others},
  journal   = {J. Affect. Disord.},
  volume    = {370},
  pages     = {260--267},
  year      = {2025}
}

@article{hall2019association,
  title     = {The association between disaster exposure and media use on post-traumatic stress disorder following Typhoon Hato in Macao, China},
  author    = {Hall, Brian J and others},
  journal   = {Eur. J. Psychotraumatol.},
  volume    = {10},
  number    = {1},
  pages     = {1558709},
  year      = {2019}
}

@article{jiang2024mixtral,
  title     = {Mixtral of experts},
  author    = {Jiang, Albert Q and others},
  journal   = {arXiv preprint arXiv:2401.04088},
  year      = {2024}
}

@misc{ivison2023camels,
  title     = {Camels in a Changing Climate: Enhancing LM Adaptation with Tulu 2}, 
  author    = {Hamish Ivison and others},
  year      = {2023},
  eprint    = {2311.10702},
  archivePrefix={arXiv},
  primaryClass={cs.CL}
}

@article{bodenreider2004unified,
  title     = {The unified medical language system (UMLS): integrating biomedical terminology},
  author    = {Bodenreider, Olivier},
  journal   = {Nucleic Acids Res.},
  volume    = {32},
  number    = {suppl\_1},
  pages     = {D267--D270},
  year      = {2004}
}

@article{alsentzer2019publicly,
  title     = {Publicly available clinical BERT embeddings},
  author    = {Alsentzer, Emily and others},
  journal   = {arXiv preprint arXiv:1904.03323},
  year      = {2019}
}

@article{ouyang2022training,
  title     = {Training language models to follow instructions with human feedback},
  author    = {Ouyang, Long and others},
  journal   = {Proc. NeurIPS},
  volume    = {35},
  pages     = {27730--27744},
  year      = {2022}
}

@article{sanh2019distilbert,
  title     = {DistilBERT, a distilled version of BERT: smaller, faster, cheaper and lighter},
  author    = {Sanh, Victor and others},
  journal   = {arXiv preprint arXiv:1910.01108},
  year      = {2019}
}

@inproceedings{liu2025pychoagent,    
  title     = {PychoAgent: Psychology-driven LLM Agents for Explainable Panic Prediction on Social Media during Sudden Disaster Events},
  author    = {Liu, Mengzhu and others},
  booktitle = {Proc. EMNLP},
  pages     = {17127--17145},
  year      = {2025}
}

@phdthesis{shafi2025wellbeingagent,
  title     = {WellbeingAgent: An LLM-Driven Agentic Framework for Personalized Mental Health Support},
  author    = {Shafi, Fozle Rabbi},
  year      = {2025}
}

@article{mclennan2020conceptualising,
  title     = {Conceptualising and measuring psychological preparedness for disaster: The Psychological Preparedness for Disaster Threat Scale},
  author    = {McLennan, Jim and others},
  journal   = {Nat. Hazards},
  volume    = {101},
  number    = {1},
  pages     = {297--307},
  year      = {2020}
}

@article{zhang2025survey,
  title     = {A survey on the memory mechanism of large language model-based agents},
  author    = {Zhang, Zeyu and others},
  journal   = {ACM TOIS},
  volume    = {43},
  number    = {6},
  pages     = {1--47},
  year      = {2025}
}

@article{cheng2024exploring,
  title     = {Exploring large language model based intelligent agents: Definitions, methods, and prospects},
  author    = {Cheng, Yuheng and others},
  journal   = {arXiv preprint arXiv:2401.03428},
  year      = {2024}
}

@article{liu2025advances,
  title     = {Advances and challenges in foundation agents: From brain-inspired intelligence to evolutionary, collaborative, and safe systems},
  author    = {Liu, Bang and others},
  journal   = {arXiv preprint arXiv:2504.01990},
  year      = {2025}
}

@article{masterman2024landscape,
  title     = {The landscape of emerging ai agent architectures for reasoning, planning, and tool calling: A survey},
  author    = {Masterman, Tula and others},
  journal   = {arXiv preprint arXiv:2404.11584},
  year      = {2024}
}

@article{huang2024understanding,
  title     = {Understanding the planning of LLM agents: A survey},
  author    = {Huang, Xu and others},
  journal   = {arXiv preprint arXiv:2402.02716},
  year      = {2024}
}

@article{zhang2022natural,
  title     = {Natural language processing applied to mental illness detection: a narrative review},
  author    = {Zhang, Tianlin and others},
  journal   = {NPJ Digit. Med.},
  volume    = {5},
  number    = {1},
  pages     = {46},
  year      = {2022}
}

@article{jiaying2025displaying,
  title     = {Displaying Fear, Sadness, and Joy in Public: Schizophrenia Vloggers' Video Narration of Emotion and Online Care-Seeking.},
  author    = {Jiaying (Lizzy) Liu and others},
  journal   = {CoRR},
  year      = {2025}
}

@article{jeong2023exploring,
  title     = {Exploring the use of natural language processing for objective assessment of disorganized speech in schizophrenia},
  author    = {Jeong, Lydia and others},
  journal   = {Psychiatry Res. Clin. Pract.},
  volume    = {5},
  number    = {3},
  pages     = {84--92},
  year      = {2023}
}

@article{plank2024reduced,
  title     = {Reduced speech coherence in psychosis-related social media forum posts},
  author    = {Plank, Laurin and Zlomuzica, Armin},
  journal   = {Schizophrenia},
  volume    = {10},
  number    = {1},
  pages     = {60},
  year      = {2024}
}

@article{guntuku2019language,
  title     = {Language of ADHD in adults on social media},
  author    = {Guntuku, Sharath Chandra and others},
  journal   = {J. Atten. Disord.},
  volume    = {23},
  number    = {12},
  pages     = {1475--1485},
  year      = {2019}
}

@article{chen2023exploring,
  title     = {Exploring the behavior of users with attention-deficit/hyperactivity disorder on Twitter: comparative analysis of tweet content and user interactions},
  author    = {Chen, Liuliu and others},
  journal   = {JMIR},
  volume    = {25},
  pages     = {e43439},
  year      = {2023}
}

@article{alsharif2024adhd,
  title     = {ADHD diagnosis using text features and predictive machine learning and deep learning algorithms},
  author    = {Alsharif, Nizar and others},
  journal   = {J. Disabil. Res.},
  volume    = {3},
  number    = {7},
  pages     = {20240082},
  year      = {2024}
}

@inproceedings{lee2024detecting,
  title     = {Detecting a proxy for potential comorbid adhd in people reporting anxiety symptoms from social media data},
  author    = {Lee, Claire and others},
  booktitle = {Proc. CLPsych},
  pages     = {172--176},
  year      = {2024}
}

@article{zhu2025leveraging,
  title     = {Leveraging large language models and traditional machine learning ensembles for ADHD detection from narrative transcripts},
  author    = {Zhu, Yuxin and others},
  journal   = {arXiv preprint arXiv:2505.21324},
  year      = {2025}
}

@article{chatzakou2019detecting,
  title     = {Detecting cyberbullying and cyberaggression in social media},
  author    = {Chatzakou, Despoina and others},
  journal   = {ACM TWEB},
  volume    = {13},
  number    = {3},
  pages     = {1--51},
  year      = {2019}
}

@article{bozyiugit2021cyberbullying,
  title     = {Cyberbullying detection: Utilizing social media features},
  author    = {Bozyi{\u{g}}it, Alican and others},
  journal   = {Expert Syst. Appl.},
  volume    = {179},
  pages     = {115001},
  year      = {2021}
}

@article{lopez2021early,
  title     = {Early detection of cyberbullying on social media networks},
  author    = {L{\'o}pez-Vizca{\'\i}no, Manuel F and others},
  journal   = {Future Gener. Comput. Syst.},
  volume    = {118},
  pages     = {219--229},
  year      = {2021}
}

@article{murshed2023faeo,
  title     = {FAEO-ECNN: cyberbullying detection in social media platforms using topic modelling and deep learning},
  author    = {Murshed, Belal Abdullah Hezam and others},
  journal   = {Multimed. Tools Appl.},
  volume    = {82},
  number    = {30},
  pages     = {46611--46650},
  year      = {2023}
}

@article{pericherla2024transformer,
  title     = {Transformer network-based word embeddings approach for autonomous cyberbullying detection},
  author    = {Pericherla, Subbaraju and Ilavarasan, E},
  journal   = {Int. J. Intell. Unmanned Syst.},
  volume    = {12},
  number    = {1},
  pages     = {154--166},
  year      = {2024}
}

@article{sihab2024cyberbullying,
  title     = {Cyberbullying detection of resource constrained language from social media using transformer-based approach},
  author    = {Sihab-Us-Sakib, Syed and others},
  journal   = {Nat. Lang. Process. J.},
  volume    = {9},
  pages     = {100104},
  year      = {2024}
}

@inproceedings{vanpech2024detecting,
  title     = {Detecting cyberbullying on social networks using language learning model},
  author    = {Vanpech, Pichapa and others},
  booktitle = {Proc. KST},
  pages     = {161--166},
  year      = {2024}
}

@article{perera2024cyberbullying,
  title     = {Cyberbullying detection system on social media using supervised machine learning},
  author    = {Perera, Andrea and Fernando, Pumudu},
  journal   = {Procedia Comput. Sci.},
  volume    = {239},
  pages     = {506--516},
  year      = {2024}
}

@inproceedings{garcia2024promoting,
  title     = {Promoting security and trust on social networks: Explainable cyberbullying detection using Large Language Models in a stream-based Machine Learning framework},
  author    = {Garc{\'\i}a-M{\'e}ndez, Silvia and De Arriba-P{\'e}rez, Francisco},
  booktitle = {Proc. SNAMS},
  pages     = {25--32},
  year      = {2024}
}

@inproceedings{lin2025ask,
  title     = {Ask, acquire, understand: A multimodal agent-based framework for social abuse detection in memes},
  author    = {Lin, Xuanrui and others},
  booktitle = {Proc. WWW},
  pages     = {4734--4744},
  year      = {2025}
}

@article{prabhu2025comprehensive,
  title     = {A comprehensive framework for multi-modal hate speech detection in social media using deep learning},
  author    = {Prabhu, R and Seethalakshmi, V},
  journal   = {Sci. Rep.},
  volume    = {15},
  number    = {1},
  pages     = {13020},
  year      = {2025}
}

@article{mali2025automatic,
  title     = {Automatic detection of cyberbullying behaviour on social media using Stacked Bi-Gru attention with BERT model},
  author    = {Mali, Mohan K and others},
  journal   = {Expert Syst. Appl.},
  volume    = {262},
  pages     = {125641},
  year      = {2025}
}

@article{jiang2025social,
  title     = {Social-llm: Modeling user behavior at scale using language models and social network data},
  author    = {Jiang, Julie and Ferrara, Emilio},
  journal   = {Sci},
  volume    = {7},
  number    = {4},
  pages     = {138},
  year      = {2025}
}

@article{shah2025advancing,
  title     = {Advancing depression detection on social media platforms through fine-tuned large language models},
  author    = {Shah, Shahid Munir and others},
  journal   = {Online Soc. Netw. Media},
  volume    = {46},
  pages     = {100311},
  year      = {2025}
}

@inproceedings{wang2025posts,
  title     = {From posts to timelines: Modeling mental health dynamics from social media timelines with hybrid llms},
  author    = {Wang, Zimu and others},
  booktitle = {Proc. CLPsych},
  pages     = {249--255},
  year      = {2025}
}

@inproceedings{wu2025psychological,
  title     = {Psychological health knowledge-enhanced LLM-based social network crisis intervention text transfer recognition method},
  author    = {Wu, Shurui and others},
  booktitle = {Proc. HBD},
  pages     = {156--161},
  year      = {2025}
}

@inproceedings{zheng2024cascade,
  title     = {Cascade Large Language Model via In-Context Learning for Depression Detection on Chinese Social Media},
  author    = {Zheng, Tong and others},
  booktitle = {Proc. PRCV},
  pages     = {353--366},
  year      = {2024}
}

@inproceedings{lashgari2025sentinel,
  title     = {SENTINEL-LLM: Suicide Ensemble-Based Text Intelligence and Natural Language Evaluation Through Large Language Models},
  author    = {Lashgari, Farzaneh and others},
  booktitle = {Proc. ICWR},
  pages     = {299--305},
  year      = {2025}
}

@inproceedings{leow2025comparison,
  title     = {Comparison of Depression Detection Between LLMs and Zero-Shot Learning Using DAD Dataset},
  author    = {Leow, Justin JD and others},
  booktitle = {Proc. CSPA},
  pages     = {295--300},
  year      = {2025}
}

@inproceedings{nguyen2025supporters,
  title     = {Supporters and Skeptics: LLM-based Analysis of Engagement with Mental Health (Mis) Information Content on Video-sharing Platforms},
  author    = {Nguyen, Viet Cuong and others},
  booktitle = {Proc. ICWSM},
  volume    = {19},
  pages     = {1329--1345},
  year      = {2025}
}

@article{qi2025supervised,
  title     = {Supervised learning and large language model benchmarks on mental health datasets: Cognitive distortions and suicidal risks in chinese social media},
  author    = {Qi, Hongzhi and others},
  journal   = {Bioengineering},
  volume    = {12},
  number    = {8},
  pages     = {882},
  year      = {2025}
}

@inproceedings{zhu2025social,
  title     = {Social Text Mental Disorder Identification Framework Integrating Generative Language Model and Psychological Feature Extraction},
  author    = {Zhu, Ya and Huang, Siqing},
  booktitle = {Proc. CAIBDA},
  pages     = {1092--1097},
  year      = {2025}
}

@article{kotov2011schizophrenia,
  title     = {Schizophrenia in the internalizing-externalizing framework: a third dimension?},
  author    = {Kotov, Roman and others},
  journal   = {Schizophr. Bull.},
  volume    = {37},
  number    = {6},
  pages     = {1168--1178},
  year      = {2011}
}

@article{myers2002ten,
  title     = {Ten-year review of rating scales. II: Scales for internalizing disorders},
  author    = {Myers, Kathleen and Winters, Nancy C},
  journal   = {JAACAP},
  volume    = {41},
  number    = {6},
  pages     = {634--659},
  year      = {2002}
}

@article{lieberman2018psychotic,
  title     = {Psychotic disorders},
  author    = {Lieberman, Jeffrey A and First, Michael B},
  journal   = {N. Engl. J. Med.},
  volume    = {379},
  number    = {3},
  pages     = {270--280},
  year      = {2018}
}

@incollection{cicchetti2014developmental,
  title     = {A developmental perspective on internalizing and externalizing disorders},
  author    = {Cicchetti, Dante and Toth, Sheree L},
  booktitle = {Internalizing and Externalizing Expressions of Dysfunction},
  pages     = {1--19},
  year      = {2014},
  publisher = {Psychology Press}
}

@article{mccutcheon2020schizophrenia,
  title     = {Schizophrenia---an overview},
  author    = {McCutcheon, Robert A and others},
  journal   = {JAMA Psychiatry},
  volume    = {77},
  number    = {2},
  pages     = {201--210},
  year      = {2020}
}

@book{apa2013diagnostic,
  title     = {Diagnostic and statistical manual of mental disorders (DSM-5)},
  author    = {{American Psychiatric Association}},
  year      = {2013},
  edition   = {5th}
}

@article{cer2018universal,
  title     = {Universal sentence encoder},
  author    = {Cer, Daniel and others},
  journal   = {arXiv preprint arXiv:1803.11175},
  year      = {2018}
}

@article{Yen2014,
  title     = {Cyberbullying among male adolescents with attention-deficit/hyperactivity disorder: prevalence, correlates, and association with poor mental health status},
  author    = {Yen, Cheng-Fang and others},
  journal   = {Psychiatry Clin. Neurosci.},
  volume    = {68},
  number    = {6},
  pages     = {420--428},
  year      = {2014}
}

@article{Liu2021,
  title     = {Perpetration of and victimization in cyberbullying and traditional bullying in adolescents with attention-deficit/hyperactivity disorder: roles of impulsivity, frustration intolerance, and hostility},
  author    = {Liu, Tzu-Li and others},
  journal   = {IJERPH},
  volume    = {18},
  number    = {13},
  pages     = {6844},
  year      = {2021}
}

@article{Pineda2024,
  title     = {Cyberbullying and problematic Internet use in adolescents with ADHD: exploring the relationship with moral disengagement and social skills},
  author    = {Pineda, D and others},
  journal   = {J. Atten. Disord.},
  year      = {2024}
}

@article{SchultzeKrumbholz2023,
  title     = {Perpetrators and victims of cyberbullying among youth with conduct disorder},
  author    = {Schultze-Krumbholz, Anja and others},
  journal   = {Eur. Child Adolesc. Psychiatry},
  year      = {2023}
}

@misc{Healthspring_CD,
  title     = {Conduct disorder},
  author    = {{HealthSpring Healthier Together}},
  year      = {2020}
}

@article{LucasMolina2021,
  title     = {The association between externalizing and internalizing problems with bullying engagement in adolescents: The mediating role of social skills},
  author    = {Lucas-Molina, Bego{\~n}a and others},
  journal   = {IJERPH},
  volume    = {18},
  number    = {19},
  pages     = {10444},
  year      = {2021}
}

@inproceedings{lewis2020bart,
  title     = {BART: Denoising sequence-to-sequence pre-training for natural language generation, translation, and comprehension},
  author    = {Lewis, Mike and others},
  booktitle = {Proc. ACL},
  pages     = {7871--7880},
  year      = {2020}
}

@article{hu2021privacy,
  title     = {Privacy data propagation and preservation in social media: A real-world case study},
  author    = {Hu, Xiangyu and others},
  journal   = {IEEE TKDE},
  volume    = {35},
  number    = {4},
  pages     = {4137--4150},
  year      = {2021}
}

@article{gao2022modeling,
  title     = {Modeling health stage development of patients with dynamic attributed graphs in online health communities},
  author    = {Gao, Yuyang and others},
  journal   = {IEEE TKDE},
  volume    = {35},
  number    = {2},
  pages     = {1831--1843},
  year      = {2022}
}

@article{wang2021context,
  title     = {How context or knowledge can benefit healthcare question answering?},
  author    = {Wang, Xiaoli and others},
  journal   = {IEEE TKDE},
  volume    = {35},
  number    = {1},
  pages     = {575--588},
  year      = {2021}
}

@inproceedings{lee2023towards,
  title     = {Towards suicide prevention from bipolar disorder with temporal symptom-aware multitask learning},
  author    = {Lee, Daeun and others},
  booktitle = {Proc. KDD},
  pages     = {4357--4369},
  year      = {2023}
}

@inproceedings{wang2023contrastive,
  title     = {Contrastive learning of stress-specific word embedding for social media based stress detection},
  author    = {Wang, Xin and others},
  booktitle = {Proc. KDD},
  pages     = {5137--5149},
  year      = {2023}
}

@article{ragheb2021negatively,
  title     = {Negatively correlated noisy learners for at-risk user detection on social networks: A study on depression, anorexia, self-harm, and suicide},
  author    = {Ragheb, others},
  journal   = {IEEE TKDE},
  volume    = {35},
  number    = {1},
  pages     = {770--783},
  year      = {2021}
}

@inproceedings{liu2024emollms,
  title     = {Emollms: A series of emotional large language models and annotation tools for comprehensive affective analysis},
  author    = {Liu, Zhiwei and others},
  booktitle = {Proc. KDD},
  pages     = {5487--5496},
  year      = {2024}
}
	 
\end{document}